\pdfoutput=1

\documentclass[11pt]{article}

\usepackage[]{acl}
\usepackage{xcolor}

\usepackage{times}
\usepackage{latexsym}

\usepackage[T1]{fontenc}

\usepackage[utf8]{inputenc}

\usepackage{microtype}

\usepackage{inconsolata}
\usepackage[ruled,vlined,noresetcount]{algorithm2e}
\usepackage{amsmath,bm}
\usepackage{inconsolata}
\usepackage{blindtext}
\usepackage{tcolorbox}
\usepackage{tabularx}

\usepackage{url}
\usepackage{multicol}
\usepackage{booktabs}
\usepackage{makecell}
\usepackage{amsmath, amssymb}
\usepackage{multirow}
\usepackage{mathtools}
\usepackage{enumitem}
\usepackage{float}
\usepackage{graphicx}
\usepackage{upgreek}
\usepackage{seqsplit}
\usepackage{color,soul}
\usepackage{arydshln}
\usepackage{placeins}
\usepackage{pifont}
\usepackage{amssymb}
\usepackage{bbm}
\usepackage{varwidth}

\newcommand{\xmark}{\textcolor{red}{\ding{55}}}

\newcommand{\hlc}[2][yellow]{ {\sethlcolor{#1} \hl{#2}} }

\usepackage{amsmath,amsfonts,bm}

\def\eqref#1{equation~\ref{#1}}

\def\1{\bm{1}}

\DeclareMathAlphabet{\mathsfit}{\encodingdefault}{\sfdefault}{m}{sl}
\SetMathAlphabet{\mathsfit}{bold}{\encodingdefault}{\sfdefault}{bx}{n}

\usepackage{caption}
\usepackage{subcaption}

\title{Typos that Broke the RAG's Back: Genetic Attack on RAG Pipeline \\ by Simulating Documents in the Wild via Low-level Perturbations}

\author{Sukmin Cho$^1$
        \quad Soyeong Jeong$^2$
        \quad Jeongyeon Seo$^1$
        \quad Taeho Hwang$^1$
        \quad Jong C. Park$^1$\thanks{\hspace{0.2cm}Corresponding author} \\
        School of Computing$^1$ \quad Graduate School of AI$^2$ \\
        Korea Advanced Institute of Science and Technology$^{1,2}$\\ 
       \texttt{\{nelllpic,starsuzi,yena.seo,doubleyyh,jongpark\}@kaist.ac.kr}}

\begin{document}
\maketitle
\begin{abstract}
The robustness of recent Large Language Models (LLMs) has become increasingly crucial as their applicability expands across various domains and real-world applications.
Retrieval-Augmented Generation (RAG) is a promising solution for addressing the limitations of LLMs, yet existing studies on the robustness of RAG often overlook the interconnected relationships between RAG components or the potential threats prevalent in real-world databases, such as minor textual errors.
In this work, we investigate two underexplored aspects when assessing the robustness of RAG: 1) vulnerability to noisy documents through low-level perturbations and 2) a holistic evaluation of RAG robustness. 
Furthermore, we introduce a novel attack method, the Genetic Attack on RAG (\textit{GARAG}), which targets these aspects.
Specifically, \textit{GARAG} is designed to reveal vulnerabilities within each component and test the overall system functionality against noisy documents. 
We validate RAG robustness by applying our \textit{GARAG} to standard QA datasets, incorporating diverse retrievers and LLMs.
The experimental results show that \textit{GARAG} consistently achieves high attack success rates. Also, it significantly devastates the performance of each component and their synergy, highlighting the substantial risk that minor textual inaccuracies pose in disrupting RAG systems in the real world. Code is available at \url{https://github.com/zomss/GARAG}.

\end{abstract}

\section{Introduction}

Large Language Models (LLMs)~\cite{GPT3, GPT4} have enabled remarkable advances in diverse Natural Language Processing (NLP) tasks, especially in Question-Answering (QA) tasks~\cite{TQA, NQ}.
Despite these advances, however, LLMs face challenges in having to adapt to ever-evolving or long-tailed knowledge due to their limited parametric memory~\cite{realtime, popqa}, resulting in a hallucination where the models generate convincing yet factually incorrect text~\cite{HaluEval}.
Retrieval-Augmented Generation (RAG)~\cite{rag} has emerged as a promising solution by utilizing a retriever to fetch enriched knowledge from external databases, thus enabling accurate, relevant, and up-to-date response generation.
Specifically, RAG has shown its superior performance across diverse knowledge-intensive tasks~\cite{rag, internetaug, adaptiverag}, leading to its integration as a core component in various real-world APIs~\cite{ToolLLM, langchain, Plugin}.
Given its extensive applications, ensuring robustness under diverse conditions of real-world scenarios becomes critical for safe deployment.
Thus, assessing potential vulnerabilities within the overall RAG system is vital, particularly by assessing its components: the retriever and the reader.

\begin{figure}[t!]
\centering
\includegraphics[width=0.9\columnwidth]{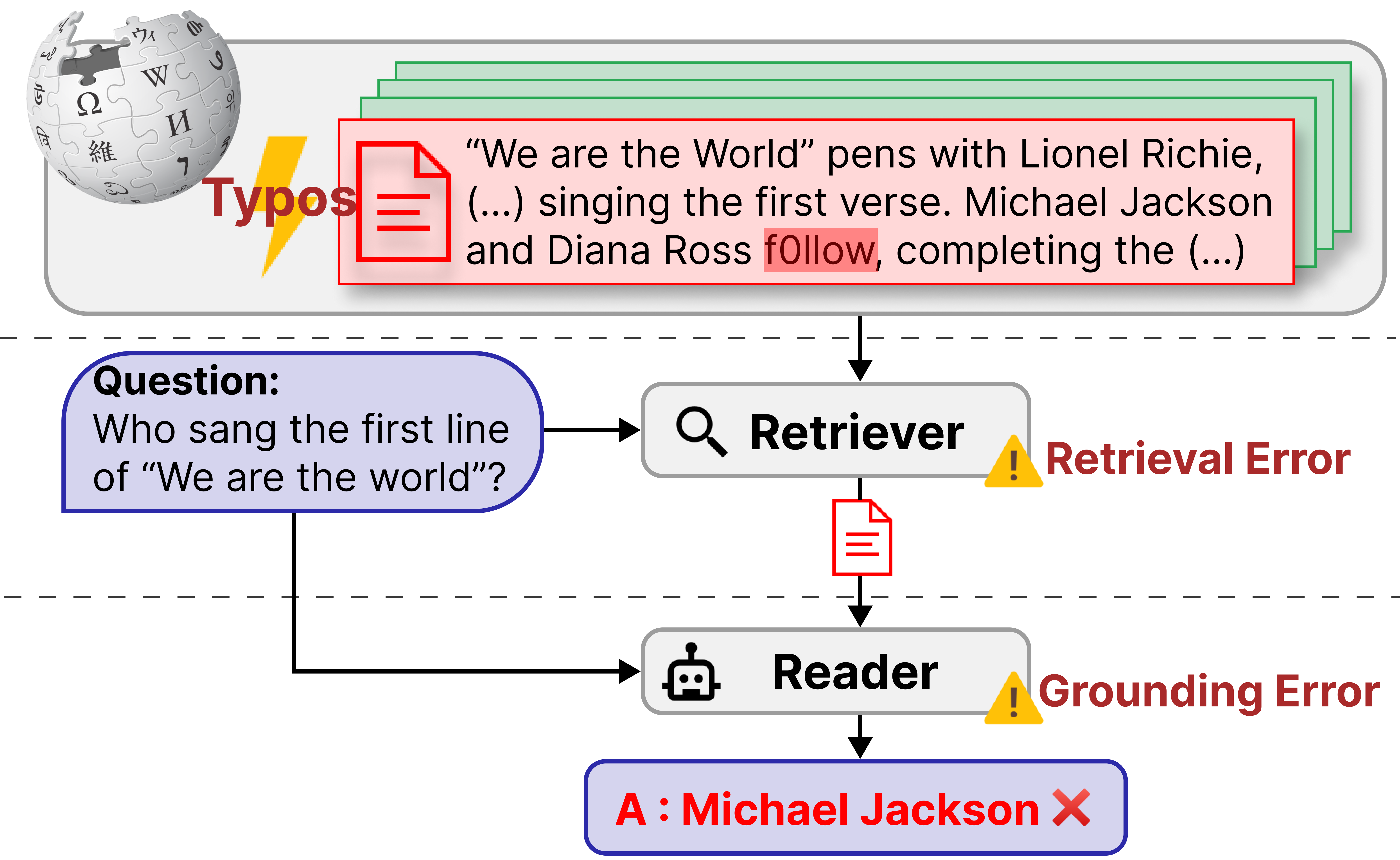}
\vspace{-.8em}
\caption{\small 
Impact of noisy documents in real-world databases on the RAG system: The retriever selects a noisy document, causing the reader to produce incorrect answers.}
\label{fig:1-1}
\vspace{-1.5em}
\end{figure}

However, existing studies on assessing the robustness of RAG often focus solely on either retrievers~\cite{PoisonRetriever, PoisonRAG, backdoor} or readers~\cite{ood, decodtrust, Promptbench}.
The robustness of a single component might only partially capture the complexities of RAG systems, where the retriever and reader work together in a sequential flow, which is crucial for optimal performance. 
In other words, the reader's ability to accurately ground information significantly depends on the retriever's capability of sourcing query-relevant documents~\cite{KALMV, ground}.
Thus, it is important to consider both components simultaneously when evaluating the robustness of an RAG system.

While concurrent work has shed light on the sequential interaction between two components, they have primarily evaluated the performance of the reader component given the high-level perturbed errors within retrieved documents, such as context relevance or counterfactual information~\cite{nomiracl, BenchmarkRAG, PoN}.
However, they have overlooked the impact of low-level errors, such as textual typos due to human mistakes or preprocessing inaccuracies in retrieval corpora, which often occur in real-world scenarios~\cite{Web_Oyster, CryptText}.
Additionally, LLMs, commonly used as readers, struggle to produce accurate predictions when confronted with textual errors~\cite{Promptbench,decodtrust}.
Note that these are the practical issues that can affect the performance of any RAG system in real-world scenarios, as illustrated in Figure~\ref{fig:1-1}.
Therefore, to deploy a more realistic RAG system, we should consider: ``\textit{Can minor document typos comprehensively disrupt both the retriever and reader components in RAG systems?}''

In this paper, we evaluate the RAG system's robustness against textual typos in the database by generating a perturbed counterpart of the clean document retrieved for a given query.
Initially, we establish two attack objectives to qualitatively measure the negative impact of the adversarial document on the RAG system's retrieval and grounding capabilities. 
To comprehensively assess system resilience under these objectives, we propose a novel black-box adversarial attack method, \textit{GARAG}, which uses a genetic algorithm to search for the most adversarial document with low values for both loss objectives among the perturbed documents.
The method begins by generating an initial population of adversarial documents by injecting minor textual errors into the original document while ensuring that answer tokens remain unaltered. 
Through an iterative process of mutation, crossover, and selection to refine the population, the method searches for the most adversarial document for a given query by effectively exploring the vast search space of typos space and exploiting the most adversarial documents.
To sum up, \textit{GARAG} assesses the holistic robustness of an RAG system against minor textual errors, offering insights into the system’s resilience through iterative adversarial refinement.

We validate our method on three standard QA datasets~\cite{TQA, NQ, SQD}, with diverse retrievers~\cite{DPR, contriever} and LLMs~\cite{Llama2, vicuna, mistral}.  
The experimental results reveal that adversarial documents with low-level perturbation generated by \textit{GARAG} significantly induce retrieval and grounding errors, achieving a high attack success rate of approximately 70\%, along with a significant reduction in the performance of each component and the overall system.
Our analyses also highlight that lower perturbation rates pose a greater threat to the RAG system, emphasizing the challenges of mitigating such inconspicuous yet critical vulnerabilities.

Our contributions in this paper are threefold:
\vspace{-0.075in}
\begin{itemize}[itemsep=0.3mm, parsep=1pt, leftmargin=*]
    \item We point out that the RAG system is vulnerable to minor but frequent textual errors within the documents, prevalent in real-world scenarios.
    \item We propose a black-box adversarial attack method, \textit{GARAG}, based on a genetic algorithm searching for adversarial documents targeting both components within RAG simultaneously.
    \item We experimentally show that \textit{GARAG} effectively attacks the RAG system with significant performance degradation, validating the vulnerability to textual typos.
\end{itemize}

\section{Related Work}

\subsection{Robustness in RAG}

The robustness of RAG, characterized by its ability to fetch and incorporate external information dynamically, has gained much attention for its critical role in real-world applications~\cite{langchain, Llamaindex, Plugin}.
However, previous studies concentrated on the robustness of individual components within RAG systems, either retriever or reader.
The vulnerability of the retriever is captured by injecting adversarial documents, specially designed to disrupt the retrieval capability, into retrieval corpora~\cite{PoisonRetriever, PoisonRAG, backdoor}.
Additionally, the robustness of LLMs, often employed as readers, has been critically examined for their resistance to out-of-distribution data and adversarial attacks~\cite{advGlue, ood, decodtrust, Promptbench}.
However, these studies overlook the sequential interaction between the retriever and reader components, thus not fully addressing the overall robustness of RAG systems.

In response, there is an emerging consensus on the need to assess the holistic robustness of RAG, with a particular emphasis on the sequential interaction of the retriever and reader~\cite{nomiracl, BenchmarkRAG}.
They point out that RAG's vulnerabilities stem from retrieval inaccuracies and inconsistencies in how the reader interprets retrieved documents. 
Specifically, the reader generates incorrect responses if the retriever fetches partially (or entirely) irrelevant or counterfactual documents within the retrieved set.
The solutions to these challenges range from prompt design~\cite{das, self-ask} and plug-in models~\cite{KALMV} to specialized language models for enhancing RAG's performance~\cite{JudgeThenGen, selfrag}.
However, they focus on the high-level errors within retrieved documents, which may overlook more subtle yet realistic low-level errors frequently encountered in the real world.

In this study, we spotlight a novel vulnerability in RAG systems related to low-level textual errors found in retrieval corpora, often originating from human mistakes or preprocessing inaccuracies~\cite{BEIR, Web_Oyster, CryptText}.
Specifically,~\citet{WikipediaTypo} pointed out that Wikipedia, a widely used retrieval corpus, frequently contains minor errors within its contents.
Therefore, we focus on a holistic evaluation of the RAG system's robustness against pervasive low-level text perturbations, emphasizing the critical need for systems that can maintain comprehensive effectiveness for real-world data.

\subsection{Adversarial Attacks in NLP}
Adversarial attacks involve generating adversarial samples designed to meet specific objectives to measure the robustness of models~\cite{adversarial_survey}.
In NLP, such attacks use a transformation function to inject perturbations into text, accompanied by a search algorithm that identifies the most effective adversarial sample.

The operations of the transformation function can be categorized into high-level and low-level perturbations. 
High-level perturbations leverage semantic understanding~\cite{genetic_1, paraphrase, textfooler}, while low-level perturbations are based on word or character-level changes, simulating frequently occurring errors~\cite{viper, zeroe, anthro, punctuation}.

Search algorithms aim to find optimal adversarial samples by identifying victim tokens in the original document, chosen based on their word importance as calculated by a single target model. 
For instance, deletion-based scoring~\cite{deletion_search} identifies important tokens by assessing increases in attack objectives when a token is deleted, while gradient-based scoring~\cite{gradient_search} uses the gradient of the attack objective for each token. 
Since these methods are unsuitable for multi-objective scenarios, a genetic algorithm that randomly selects tokens with elaborate exploitation is more effective~\cite{genetic_1, genetic_2, genetic_3}. 
To evaluate the robustness of the overall RAG system, which has non-differentiable and dual objectives for a retriever and a reader, we propose a novel attack algorithm incorporating a genetic algorithm.



\section{Method}

Here, we introduce our problem formulation and a novel attack method, \textit{GARAG}. Further details of the proposed method are described in Appendix~\ref{sup:operation}.

\subsection{Problem Formulation}

\paragraph{Pipeline of RAG.} 
Let \(\bm{q}\) be a query the user requests. 
In a RAG system, the retriever first fetches the query-relevant document \(\bm{d}\), then the reader generates the answer grounded on document-query pair \((\bm{d},\bm{q})\).
The retriever, parameterized with \(\phi=(\phi_d,\phi_q)\), identifies the most relevant document in the database. 
The relevance score \(r\) is computed by the dot product of the embeddings for document \(\bm{d}\) and query \(\bm{q}\), as \(r_\phi(\bm{d},\bm{q}) = \texttt{Enc}(\bm{d};\phi_d) \cdot \texttt{Enc}(\bm{q};\phi_q)\).
Finally, the reader, using an LLM parameterized with \(\theta\), generates the answer \(\bm{a}\) from the document-query pair (\(\bm{d},\bm{q}\)), as \(\bm{a} = \texttt{LLM}(\bm{d}, \bm{q}; \theta)\).

\paragraph{Adversarial Document Generation.}
To simulate the adversarial document having typical noise encountered in real-world scenarios, we introduce low-level perturbations to mimic these conditions.
We generate an adversarial document \(\bm{d'}\) by transforming the clean document \(\bm{d}\) using a function \(f\) that alters each token \(d\) into a perturbed version \(d'\). 
The function \(f\) randomly applies one of several operations — inner-shuffling, truncation, keyboard errors, or natural typos — to each token, then outputs the perturbed token: \(d'=f(d)\).
This randomness reflects the unpredictable nature of textual typos. 
Therefore, we explore a broad search space of potential adversarial documents generated from \(\bm{d}\) using \(f\) to identify the adversarial document for the RAG system,


\paragraph{Attack Objective on RAG.}
To identify an adversarial document \(\bm{d'}\) that challenges the capabilities of the RAG, we compare its negative impact against the original document \(\bm{d}\) for a given query \(\bm{q}\). 
The goal is for \(\bm{d'}\) to divert attention from \(\bm{d}\), ensuring that \(\bm{d}\) no longer appears as the top result for \(\bm{q}\). Additionally, \(\bm{d'}\) should mislead LLM into generating an incorrect answer \(\bm{a'}\) when paired with \((\bm{d^*}, \bm{q})\). To measure this negative impact, we use two loss objectives: the Relevance Score Ratio (RSR) and the Generation Probability Ratio (GPR) for retrieval and grounding, respectively.


The RSR calculates the ratio of the relevance score\footnote{Given the potential for relevance scores to be negative, we have structured the term to guarantee positivity.} from the adversarial document \(\bm{d'}\) to the score from the original document \(\bm{d}\) for the given query \(\bm{q}\). Conversely, the GPR calculates the ratio of the generation probability\footnote{The generation probability represents the joint probabilities over the answer tokens given a single document and a single question.} of the correct answer \(\bm{a}\) from the original pair \((\bm{d}, \bm{q})\) to the probability from the adversarial pair \((\bm{d'}, \bm{q})\). These two metrics are formally represented as:
\begin{equation}
    \mathcal{L}_\textnormal{RSR}(\bm{d'}) = \frac{e^{r_\phi(\bm{d},\bm{q})}}{e^{r_\phi(\bm{d'},\bm{q})}}, \mathcal{L}_\textnormal{GPR}(\bm{d'}) = \frac{p_\theta(\bm{a}|\bm{d'},\bm{q})}{p_\theta(\bm{a}|\bm{d},\bm{q})}.
\end{equation}
The values below 1 signify that a noisy document \(\bm{d'}\) generated from the adversarial attack successfully satisfies the attack objectives of distracting the retriever and misleading LLM.
Note that, as these objectives are designed for adversarial attacks, they don't directly align with each module's performance measured by conventional metrics.

Consequently, the search for an optimal adversarial document within the RAG system is defined as a dual-objective optimization problem, aiming to minimize both the RSR and GPR simultaneously:
\begin{equation}
    \bm{d^*}=\underset{\bm{d'}\in D'}{\arg\min} (\mathcal{L}_\text{RSR}(\bm{d'}), \mathcal{L}_\text{GPR}(\bm{d'}))
\label{Eq3}
\end{equation}
This optimization problem involves dual-model environments, resulting in non-differentiable conditions. To design effective adversarial attack methods targeting the RAG system through noisy document simulation, these methods must address the challenges of dual-objective and dual-model optimization within a vast search space characterized by unpredictable and diverse textual typos.


\subsection{GARAG: Genetic Attack on RAG}
\captionsetup[figure*]{skip=5pt}
\begin{figure*}[t!]
\centering
\includegraphics[width=0.85\textwidth]{Fig/Frame_2_6.png}
\caption{\small (Left) The search space formulated by our proposed attack objectives, \(\mathcal{L}_{\textnormal{RSR}}\) and \(\mathcal{L}_{\textnormal{GPR}}\). (Right) An overview of the iterative process implemented by our proposed method, \textit{GARAG}.}
\label{fig:3-1}
\vspace{-1.0em}
\end{figure*}


In this work, we introduce a novel black-box adversarial attack method called \textit{GARAG}, employing a genetic algorithm to address the dual-objective and dual-model optimization problem in a large search space. 
Initially, as shown in Figure~\ref{fig:3-1}, we divide the search space into four zones based on the attack objectives: safety, retrieval error, grounding error, and holistic error. 
The adversarial document should ideally be in a holistic error zone, where retrieval and grounding errors intersect, and should be closer to the origin, indicating a more significant negative impact on the RAG system.
Then, our proposed method, \textit{GARAG}, iteratively refines a population of adversarial documents, methodically moving them closer to the origin. 
This process involves exploring the search space to discover new adversarial documents and exploit the most adversarial ones with crossover, mutation, and selection steps. 

Formally, given the query-document pair \((\bm{q},\bm{d})\) where the document \(\bm{d}=\{d_i\}_{i=1}^N\) is retrieved for the query \(\bm{q}\), our objective is to generate the adversarial counterpart \(\bm{d'}\) with \(N\cdot pr_{pert}\) perturbed tokens, where \(pr_{pert}\) is a pre-defined hyperparameter and \(N\) is the number of tokens in \(\bm{d}\). The steps, including crossover, mutation, and selection, are repeated \(N_{iter}\) times after initialization.

\paragraph{Initialization.}
Our attack begins with the initialization step. 
We first construct the initial population \(P_0\), consisting of adversarial documents \(\bm{d'}_{i}\), formalized as \(P=\{\bm{d'}_{i}\}_{i=1}^S\), where \(S\) is the total number of documents in the population.
In detail, generating the adversarial document \(\bm{d'}_{i}\) involves selecting tokens for the attack, applying perturbations, and assembling the modified document. 
Initially, to determine which tokens to alter, a subset of indices \(I'\) containing \(N \cdot pr_{\textnormal{pert.}}\) indices is randomly selected from the complete set of token indices \(I = \{1, \ldots, N\}\), where \(N\) represents the total number of tokens in the document \(\bm{d}\).
This selection is designed to exclude any indices that correspond to the correct answer \(\bm{a}\) within the document, thus ensuring that the perturbations focus exclusively on assessing the impact of noise.
Each selected token \(d_i\) is then transformed using the function \(f\), yielding a perturbed version \(d'_i\), for \(i \in I' \subset I\). 
The final document \(\bm{d'}\) merges the set of unaltered tokens \(T=\{d_i | i \notin I \setminus I'\}\) with the set of modified tokens, represented by \(T'=\{d'_j | j \in I'\}\), forming \(\bm{d'}=T \cup T'\).
In Figure~\ref{fig:3-1}, the figure on the right shows the initialization step where the initial (parent) documents are represented as orange-colored dots, given the star-shaped original document.

\paragraph{Crossover \& Mutation.}
Then, through the crossover and mutation steps, the adversarial documents are generated by balancing the exploitation of existing knowledge within the current population (parent documents) and the exploration of new documents (offspring documents).
In detail, the crossover step generates offspring documents by recombining tokens from pairs of parent documents, incorporating their most effective adversarial features.
Subsequently, the mutation step introduces new perturbations to some tokens in the offspring, aiming to explore genetic variations that are not present in the parent documents.

Formally, the crossover step selects \(N_{\textnormal{parents}}\) pairs of parent documents from the population \(P\).
Let \(\bm{d'}_0\) and \(\bm{d'}_1\) be the selected parent documents along with their perturbed token sets \(T'_0\) and \(T'_1\), respectively.
Then, the swapping tokens perturbed in each parent document generate the offspring documents, excluding those in the shared set \(T'_0 \cap T'_1\).
The number of swapping tokens is determined by the predefined crossover rate \(pr_{\textnormal{cross}}\), applied to the number of unique perturbed tokens in each document.

The mutation step selects two corresponding subsets of tokens, \(M\) from the original token set \(T\) and \(M'\) from the perturbed token set \(T'\), ensuring that both subsets are of equal size \(|M| = |M'|\). 
The size of these subsets is determined by the predefined mutation probability \(pr_{\textnormal{mut.}}\), which is applied to \(pr_{\textnormal{pert.}} \cdot N\). 
Tokens \(d_i \in M\) are altered using a perturbation function \(f\), whereas tokens \(d'_j \in M'\) are reverted to their original states \(d_j\). 
Following this, the sets of unperturbed and perturbed tokens, \(T_{\textnormal{new}}\) and \(T'_{\textnormal{new}}\), respectively, are updated to incorporate these modifications: \(T_{\textnormal{new}} = (T \setminus M) \cup M'\) and \(T'_{\textnormal{new}} = (T' \setminus M') \cup M\).
The newly mutated document, \(\bm{d'}_{\textnormal{new}}\), is composed of the updated sets \(T_{\textnormal{new}}\) and \(T'_{\textnormal{new}}\), and the offspring set \(O\) is then formed, comprising these mutated documents.
The offspring documents are represented by blue-colored dots in the figure on the right in Figure~\ref{fig:3-1}.

\paragraph{Selection.}
The remaining step is to select the most optimal adversarial documents from the combined set \(\hat{P}=P \cup O\), which includes both parent and offspring documents.
Specifically, each document within \(\hat{P}\) is evaluated against the two attack objectives, \(\mathcal{L}_{\textnormal{RSR}}\) and \(\mathcal{L}_{\textnormal{GPR}}\), to assess their effectiveness in the adversarial context.
Therefore, we incorporate a non-dominated sorting strategy~\cite{NSGA} to identify the optimal set of documents, known as the Pareto front. 
In this front, each document is characterized by having all objective values lower than those in any other set, as shown in the right of Figure~\ref{fig:3-1}. 
Then, the documents in the Pareto front will be located in a holistic error zone closer to the origin.
Additionally, to help preserve diversity within the document population, we further utilize the crowding distance sorting strategy to identify adversarial documents that possess unique knowledge by measuring how isolated each document is relative to others.
Then, the most adversarial document \(\bm{d}^*\) is selected from a less crowded region of the Pareto front.
Details of a non-dominated sorting algorithm are described in Appendix~\ref{SA}.

Note that this process, including crossover, mutation, and selection steps, continues iteratively until a successful attack is achieved, where the selected adversarial document \(\bm{d}^*\) prompts an incorrect answer $a'$, as illustrated in the figure on the right in Figure~\ref{fig:3-1}.
If the process fails to produce a successful attack, it persists through the predefined number of iterations, \(N_{\textnormal{iter.}}\).

\begin{table}[t]
\caption{\small Results of adversarial attacks using \textit{GARAG}, averaged across three datasets, NQ, TQA, and SQuAD.
The most vulnerable results are in \textbf{bold}.}
\vspace*{-1mm}
\centering
    \small
    \renewcommand{\arraystretch}{1}
\resizebox{\columnwidth}{!}{
        \begin{tabular}{ll ccc cc
        }
        \toprule
        & & \multicolumn{3}{c}{\textbf{Attack Success Ratio ($\uparrow$)}} & \multicolumn{2}{c}{\textbf{End-to-End ($\downarrow$)}}    \\ 
        \cmidrule(l{2pt}r{2pt}){3-5}  \cmidrule(l{2pt}r{2pt}){6-7}
        \textbf{Retriever} & \textbf{LLM} & ASR$_R$ & ASR$_L$ & ASR$_T$ & \;\;EM\; & \;Acc\;\; \\ \midrule
        \multirow{6}*{\textbf{DPR}} & \textbf{Llama2-7b} & 79.2 & 90.5 & 70.1 & 77.1 & 81.3   \\
        & \textbf{Llama2-13b} & 78.4 & \textbf{92.0} & 70.8 &  81.9 & 87.3 \\ \noalign{\vskip 0.5ex}\cdashline{2-7}\noalign{\vskip 0.5ex}
        & \textbf{Vicuna-7b} & 88.7 & 80.7 & 69.8 & 57.2 & 79.3 \\
        & \textbf{Vicuna-13b} & 88.8 & 81.6 & 70.8 & 58.4 & 83.2 \\
        \noalign{\vskip 0.5ex}\cdashline{2-7}\noalign{\vskip 0.5ex}
        & \textbf{Mistral-7b}  &  83.7 & 85.5 & 69.5 & 66.7 & 96.5 \\ \midrule
        \multirow{6}*{\textbf{Contriever}} & \textbf{Llama2-7b} & 85.3 & 91.0 & \textbf{76.6} & 75.0 & 79.6 \\
        & \textbf{Llama2-13b} & 82.0 & \textbf{92.0} & 74.2 & 80.7 & 87.3 \\ \noalign{\vskip 0.5ex}\cdashline{2-7}\noalign{\vskip 0.5ex}
        & \textbf{Vicuna-7b}  & \textbf{92.1} & 81.5 & 73.9 & 55.1 & \textbf{76.9}  \\
        & \textbf{Vicuna-13b} & 91.3 & 83.2 & 74.7 & \textbf{53.5} & 79.5   \\
        \noalign{\vskip 0.5ex}\cdashline{2-7}\noalign{\vskip 0.5ex}
        & \textbf{Mistral-7b} & 89.2 & 86.6 & 75.9 & 63.1 & 95.3  \\  \noalign{\vskip 0.5ex}\cdashline{1-7}\noalign{\vskip 0.5ex}
        \textbf{w/o \textit{GARAG}}  &  & - & - & - & 100 & 100  \\ 
        \bottomrule
        \end{tabular}
}
\label{tab:1}
\vspace{-.5em}
\end{table}

\section{Experimental Setup}

In this section, we describe the experimental setup.

\subsection{Model}

\paragraph{Retriever.} 
We use two recent dense retrievers: \textbf{DPR}~\cite{DPR}, a supervised one trained on query-document pairs, and \textbf{Contriever}~\cite{contriever}, an unsupervised one.

\paragraph{Reader.}
Following concurrent work~\cite{selfrag, REAR} that utilizes LLMs as readers for the RAG system, with parameters ranging from 7B to 13B, we have selected open-source LLMs of similar capacities: 
\textbf{Llama2}~\cite{Llama2}, \textbf{Vicuna}~\cite{vicuna}, and \textbf{Mistral}~\cite{mistral}. 
Each model has been either chat-versioned or instruction-tuned. 
To adapt these models for open-domain QA tasks, we employ a zero-shot prompting template for exact match QA derived from~\citet{REAR}.

\subsection{Dataset}

We leverage three representative QA datasets: \textbf{Natural Questions (NQ)}~\cite{NQ}, \textbf{TriviaQA (TQA)}~\cite{TQA}, and \textbf{SQuAD (SQD)}~\cite{SQD}, following the setups of~\citet{DPR}. 
To assess the robustness of the RAG system, we randomly extract 1,000 instances of the triple \((\bm{q}, \bm{d}, \bm{a})\).
In each triple, \(\bm{q}\) is a question from the datasets, \(\bm{d}\) is a document from the top-100 documents retrieved from the Wikipedia corpus corresponding to \(\bm{q}\), and \(\bm{a}\) is the answer generated by the LLM, which is considered as correct for the specific question-document pair. 

\subsection{Evaluation Metric}
To measure the effectiveness of \textit{GARAG} and the actual impact of generated adversarial documents on RAG systems, we incorporate two types of metrics to show the effectiveness of the adversarial attacks and the end-to-end QA performance measuring the actual impact on the RAG system.

\paragraph{Attack Success Ratio (ASR).}
Attack Success Ratio (ASR) is the ratio of the generated documents from the adversarial attack, located in the holistic error zone (i.e., the values below 1 for \(\mathcal{L}_{\textnormal{RSR}}\) and \(\mathcal{L}_{\textnormal{GPR}}\)).
Specifically, ASR is for measuring the effectiveness of the proposed method addressing dual-objective optimization problems.

\begin{table}[t]
\centering
\caption{\small Retrieval performance under RAG system using Llama-7b when the adversarial documents generated by \textit{GARAG} are injected into the retrieval corpus.}
\vspace*{-1mm}
\label{tab:2}
\renewcommand{\arraystretch}{1.0}
\resizebox{\columnwidth}{!}{
\begin{tabular}{c c ccc ccc}
\toprule   & & \multicolumn{3}{c}{\textbf{DPR}} & \multicolumn{3}{c}{\textbf{Contriever}} \\ 
\cmidrule(l{2pt}r{2pt}){3-5} \cmidrule(l{2pt}r{2pt}){6-8} 
 \textbf{Dataset} &  \textbf{Attacked} & \scriptsize{MAP@100} & \scriptsize{NDCG@100} &  \scriptsize{ASR$_R$} & \scriptsize{MAP@100} & \scriptsize{NDCG@100} &  \scriptsize{ASR$_R$} \\ 
\midrule \multirow{2}*{\textbf{NQ}} & \xmark & .417 & .633  & - & .248 & .489 & - \\
& \textcolor{green}{\checkmark} & .356 & .593 & 75.4 & .219 & .462 & 85.9 \\ 
\midrule \multirow{2}*{\textbf{TQA}} & \xmark & .532 & .740  & - & .337 & .696 & - \\
& \textcolor{green}{\checkmark} & .471 & .696 & 78.2 & .298 & .559 & 84.9 \\
\midrule \multirow{2}*{\textbf{SQD}} & \xmark & .321 & .540  & - & .267 & .498 & - \\
& \textcolor{green}{\checkmark} & .279 & .513 & 80.0 & .223 & .468 & 86.1 \\ \bottomrule
\end{tabular}
}
\vspace{-.5em}
\end{table}


\paragraph{End-to-End Performance (E2E).} 
To evaluate the impact of the adversarial document on RAG systems, we report it with standard QA metrics: \textbf{Exact Match (EM)} and \textbf{Accuracy (Acc)}.
EM evaluates if a prediction precisely matches the correct answer, while Acc checks if the answer span is included in the predicted response. 
If the attack fails (i.e., either value for \(\mathcal{L}_{\textnormal{RSR}}\) or \(\mathcal{L}_{\textnormal{GPR}}\) exceeds 1), we transmit the original document \(\bm{d}\) to LLM instead of the adversarial one \(\bm{d'}\) during prediction.

\subsection{Implementation Details}

The proposed method, \textit{GARAG}, was configured with hyperparameters: \(N_{\textnormal{iter}}\) was set to 25, \(N_{\textnormal{parents}}\) to 10, and \(S\) to 25. \(pr_\textnormal{pert}\), \(pr_\textnormal{cross}\), and \(pr_\textnormal{mut}\) were set to 0.2, 0.2, and 0.4, respectively. The operations of perturbation function $f$ in \textit{GARAG} consist of the inner swap, truncate, keyboard typo, and natural typo, following \citet{zeroe}\footnote{\url{https://github.com/yannikbenz/zeroe}}. For computing resources, we use A100 GPU clusters.

\begin{figure*}[t!]
    \vspace{-.5em}
    \begin{minipage}[t]{0.66\textwidth}
        \begin{minipage}[t]{0.5\textwidth}
            \centering
            \includegraphics[width=0.99\columnwidth]{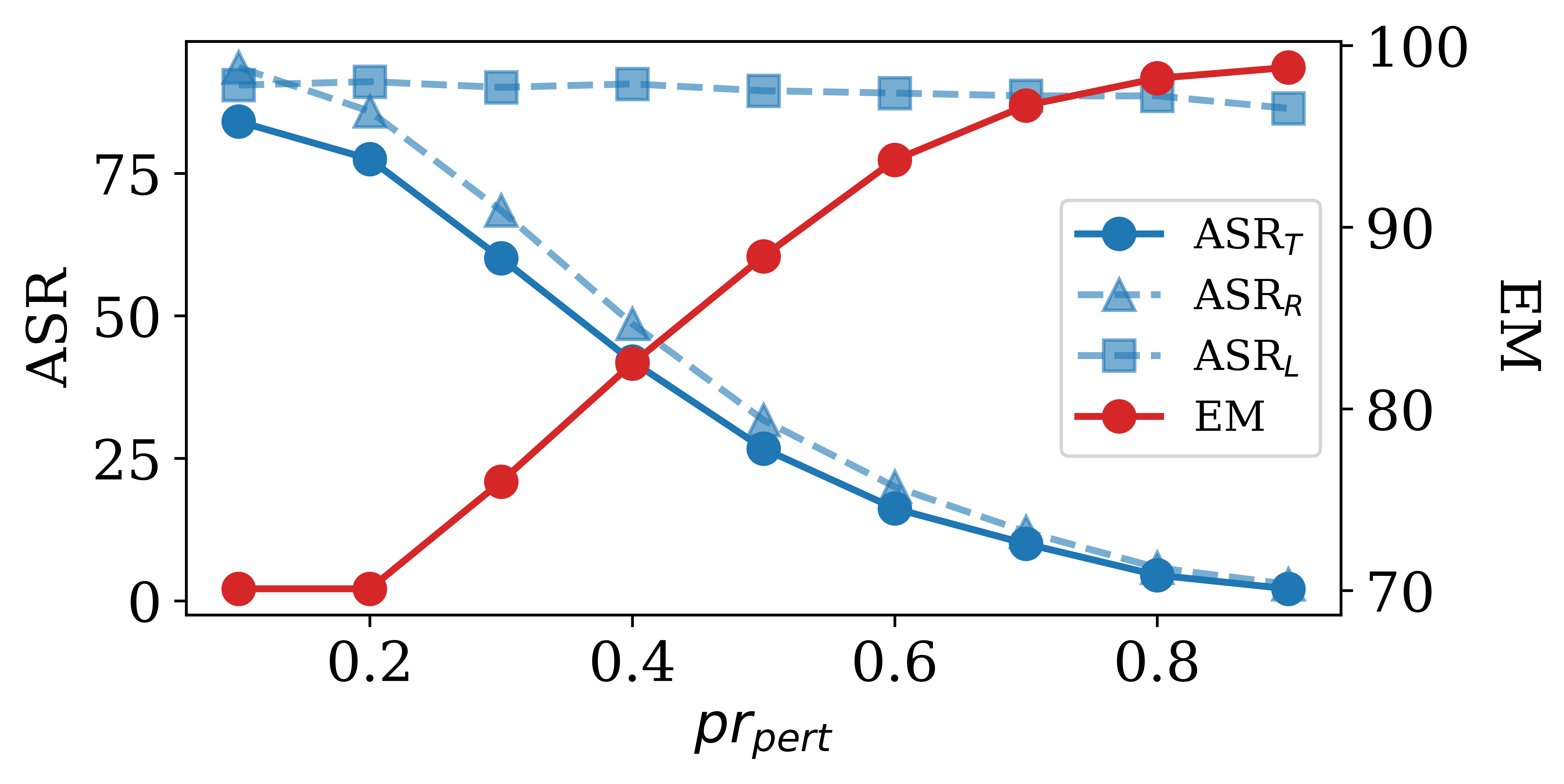}
        \end{minipage}
        \hfill
        \begin{minipage}[t]{0.5\textwidth}
            \centering
            \includegraphics[width=0.99\columnwidth]{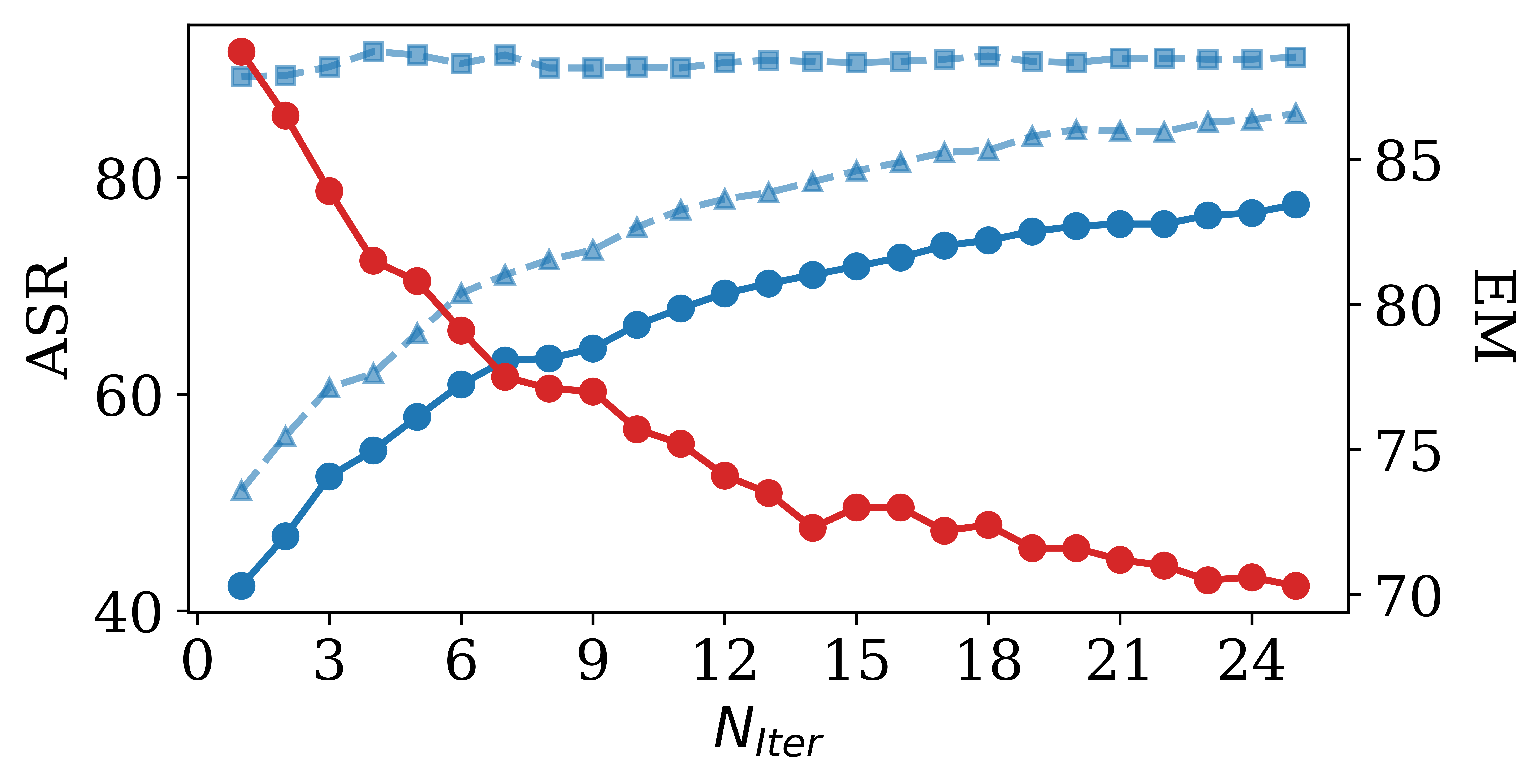}
        \end{minipage}
        \vspace*{-7mm}
    \label{fig:3}
    \end{minipage}
    \hfill
    \begin{minipage}[t]{0.327\textwidth}
        \centering
        \includegraphics[width=0.99\columnwidth]{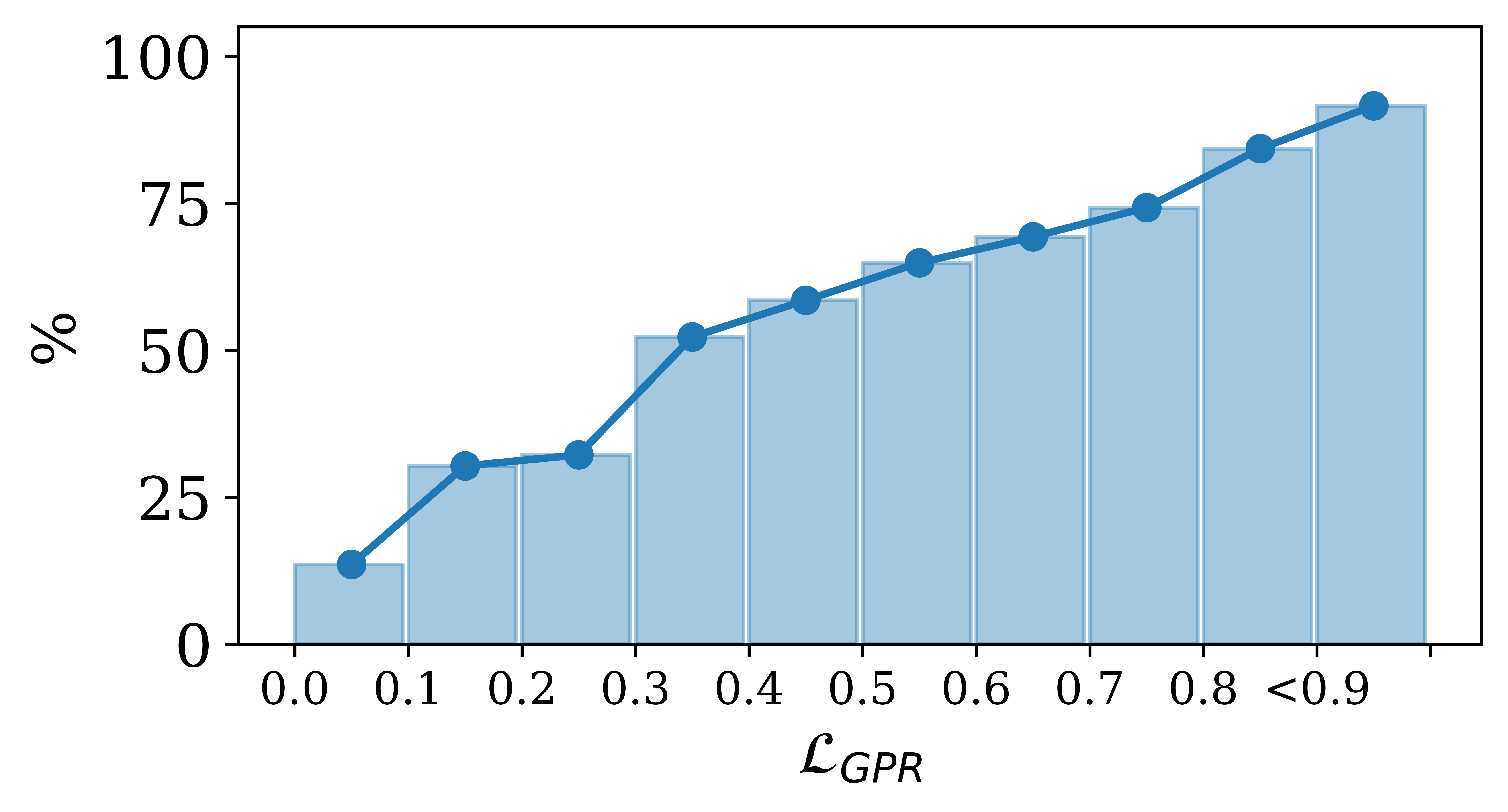}
        \vspace*{-7mm}
        \label{fig:4}
    \end{minipage}
    \caption{\small
    Adversarial attack analysis on the NQ dataset using Contriever and Llama2-7b: (Left) Variations in ASR and EM scores as the \(pr_{\textnormal{pert}}\) increases from 0 to 0.9, with ASR shown in blue and EM in red. (Center) Variations in ASR and EM scores across increasing iterations (\(N_{\textnormal{iter}}\)), also indicated in blue and red respectively. (Right) Distribution of correctness among predictions depending on \(\mathcal{L}_{\textnormal{GPR}}\).}
    \label{fig:3}
    \vspace{-1em}
\end{figure*}



\begin{figure}[t!]
\centering
\includegraphics[width=0.85\columnwidth]{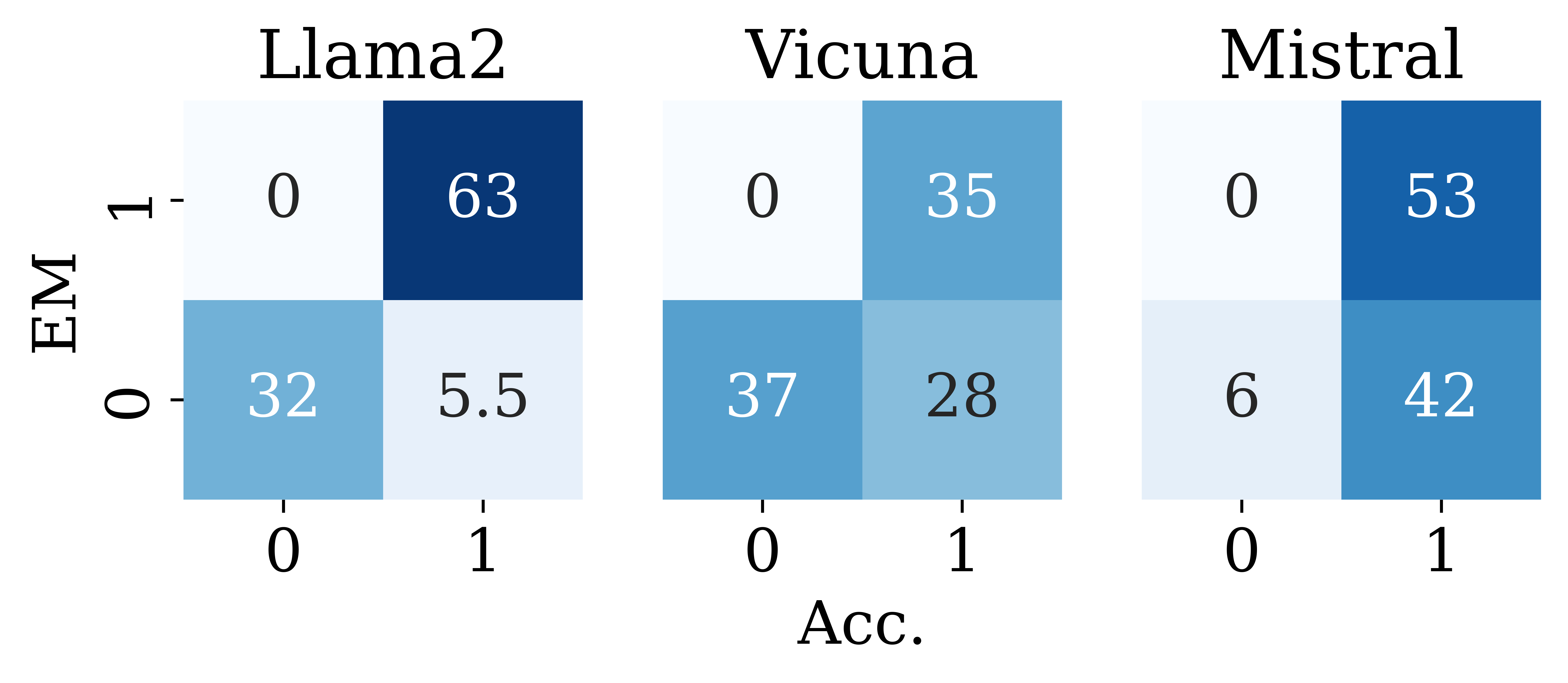}
\vspace{-.5em}
\caption{\small Confusion matrices of prediction from \(\bm{d^*}\) across EM and Acc. on NQ with Contriever.}
\label{fig:4}
    \vspace{-1em}
\end{figure}

\section{Results}

In this section, we show our experimental results with an in-depth analysis of the adversarial attack.

\subsection{Main Result.} 
Table~\ref{tab:1} shows our main results averaged over three datasets using \textit{GARAG} with two metrics: attack success ratio (ASR) and end-to-end performance (E2E). First, a notable success rate of over 70\% across all scenarios indicates that \textit{GARAG} effectively locates adversarial documents within the holistic error zone by simultaneously considering retrieval and reader errors. Additionally, we analyze the E2E performance to assess how adversarial attacks impact overall QA performance. Based on the EM metric, the performance of RAG systems decreased by an average of 30\% and a maximum of close to 50\% in all cases. These findings imply that noisy documents with minor errors, frequently found in the real world, can pose significant risks to downstream tasks using RAG.

\paragraph{Impact on Retrieval Ability.}
We qualitatively explored the impact of adversarial documents on the RAG system's retrieval ability. After injecting these documents into the original retrieval corpus, we evaluated the results using conventional IR metrics like MAP and NDCG.
As shown in Table~\ref{tab:2}, the adversarial documents degrade retrieval performance across all scenarios, despite being assessed solely by the \(\mathcal{L}_{\textnormal{RSR}}\) in the \textit{GARAG} process without considering the entire retriever corpus. Additionally, as DPR achieves better retrieval performance both before and after the attack, these results suggest that retrievers with superior retrieval performance tend to be more robust against typos.

\paragraph{Impact on Grounding Ability.} We further analyze the response patterns of LLM to adversarial documents, categorizing the results based on EM and Acc as shown in Figure~\ref{fig:4}. 
For instance, an EM of 0 and Acc of 1 indicates that the response includes the correct answer along with irrelevant tokens, whereas an EM and Acc of 0 means that the response is entirely incorrect, likely a hallucination. 
First, Llama2 tends to produce exact matches more frequently, as evidenced by a high rate of (1,1) outcomes. but struggles with completely incorrect responses under adversarial conditions, indicated by a lower proportion of (0,1). By contrast, 
Mistral, despite fewer exact matches, consistently includes the correct answer span in its responses. 
These insights are vital for understanding how different models perform in realistic scenarios, especially when handling noisy or adversarially altered documents, highlighting the varied impacts of such conditions on LLMs.


\paragraph{Impact of \(pr_{\textnormal{pert}}\) and \(N_{\textnormal{iter}}\)} Then, we further explore how varying the perturbation probability \(pr_{\textnormal{pert}}\) or the number of iterations \(N_{\textnormal{iter}}\) affects the attack outcomes. 
As the left and center figures of Figure~\ref{fig:3} illustrate, there is an apparent correlation between the attack success rates for the retriever (ASR$_R$) and the entire pipeline (ASR$_T$). 
Moreover, the consistently high success rate for the LLM (ASR$_L$) across all cases highlights a significant vulnerability in the reader against typos. 
These findings highlight the critical role of the retriever as a first line of defense in the RAG system. 
Interestingly, in the left figure of Figure~\ref{fig:3}, the results indicate that a lower proportion of perturbation within a document leads to a more disruptive impact on the RAG system. 
These experimental results suggests that documents with a few typos, which are common in the wild, could have a more detrimental effect on performance.

This phenomenon is counter-intuitive, as other attack approaches typically show that more attack vectors lead to stronger adversarial effects. 
We speculate that this occurs because the training data for neural retrievers generally consists of clean documents without typos, making it easier for the retriever to identify and reject documents with many errors. 
In contrast, LLMs, which are also trained on clean texts, struggle to generate correct answers when typos are present. The errors cause the document to lose key information or clarity, making it difficult for the model to infer the correct answer.

\paragraph{Impact of Lowering $\mathcal{L}_{\textnormal{GPR}}$.} Since the value of \(\mathcal{L}_\textnormal{GPR}\) does not directly indicate the likelihood of generating incorrect answers with auto-regressive models, we analyze the correlation between the likelihood of generating incorrect answers and \(\mathcal{L}_{\textnormal{GPR}}\).
As illustrated in the right panel of Figure~\ref{fig:3}, we categorize predictions into buckets based on their \(\mathcal{L}_{\textnormal{GPR}}\) ranges and calculate the proportion of incorrect answers within each bucket. The results validate our objective design, demonstrating that a lower \(\mathcal{L}_{\textnormal{GPR}}\) value is associated with a higher likelihood of incorrect responses.

\subsection{Analysis}

\begin{table}[t]
\centering
\small
\caption{\small Ablation study of \textit{GARAG} on NQ with Contriever and Llama-7b.}
\vspace{-.5em}
\centering
\resizebox{.85\columnwidth}{!}{
\begin{tabular}{l ccc c}
\toprule
 & \multicolumn{3}{c}{\textbf{ASR}} & \textbf{E2E}  \\ \cmidrule(l{2pt}r{2pt}){2-4} \cmidrule(l{2pt}r{2pt}){5-5} 
  & ASR$_R$ & ASR$_L$ & ASR$_T$ & EM \\ \midrule
\textbf{\textit{GARAG}} & 85.9 & 91.1 & 77.5 & 70.1  \\ \midrule
\multicolumn{5}{c}{\textit{Low-level Perturbations included \(f\)}} \\ \midrule
\textbf{Natural Typo} & 88.8 & 90.0 & 78.8 & 75.4  \\ 
\textbf{Keyboard Typo} & 84.6 & 91.4 & 76.2 & 71.2 \\ 
\textbf{Truncate} & 89.2 & 90.2 & 79.4 & 71.4 \\ 
\textbf{Inner Swap} & 83.4 & 87.8 & 71.4 & 78.0 \\ \midrule \multicolumn{5}{c}{\textit{Low-level Perturbations not included \(f\)}} \\  \midrule
\textbf{Punc.} & 93.0 & 93.7 & 86.7  & 68.9  \\ 
\textbf{Phonetic.} & 84.7 & 92.1 & 76.8  & 70.0 \\ 
\textbf{Visual.} & 77.7 & 90.5 & 68.8  & 72.5 \\ 
\bottomrule
\end{tabular}
}
\label{tab:4}
\vspace{-1.0em}
\end{table}

\begin{table}[t!]
\centering
\small
\caption{\small Adversarial attack on paraphrased query on NQ with Contriever and Llama-7b.}
\vspace{-.5em}
\centering
\renewcommand{\arraystretch}{1.1}
\resizebox{\columnwidth}{!}{
\begin{tabular}{c c cccc}
\toprule
 \textbf{Paraphrased} & \textbf{Attacked} & ASR$_R$ & ASR$_L$ & ASR$_T$ & EM \\ \midrule
 \xmark & \xmark & - & - & - & 100 \\ 
\xmark & \textcolor{green}{\checkmark} & 85.9 & 91.1 & 77.5 & 70.1 \\ 
 \noalign{\vskip 0.5ex}\cdashline{1-6}\noalign{\vskip 0.5ex}
\textcolor{green}{\checkmark} & \xmark & - & - & - & 79.1 \\ 
\textcolor{green}{\checkmark} & \textcolor{green}{\checkmark} & 72.8 & 62.5 & 44.1 & 75.1 \\ 
\bottomrule
\end{tabular}
}
\label{tab:9-3}
\vspace{-1.0em}
\end{table}



\noindent \textbf{Evaluation on Paraphrased Query.}
To create a more realistic scenario, we tested the effect of noisy documents with paraphrased queries that were not used in the adversarial attack. After generating an adversarial document for a given document-query pair, we paraphrased the query using GPT-3.5~\cite{GPT3}. These paraphrased queries, while not part of the adversarial document generation, still seek the same answers as the original ones.
As shown in Table~\ref{tab:9-3}, our results demonstrate the robustness of adversarial documents generated by \textit{GARAG}. While these documents are less effective against paraphrased queries, resulting in lower ASR and higher EM scores, they still degrade RAG system performance after attacks. The paraphrased queries also destabilize RAG systems, underscoring their vulnerability in dynamic, real-world settings like human-RAG system interactions.

\paragraph{Types of Low-level Perturbation.} 
Table~\ref{tab:4} presents the results of an ablation study on the operations included and excluded in the transformation function \(f\). 
Using multiple operations in \(f\) as the default setup consistently outperformed all single operations included in \(f\), highlighting \textit{GARAG}'s ability to exploit promising areas in a vast search space. 
Furthermore, the other types of low-level perturbations not initially included in \(f\)—such as punctuation insertion, phonetic similarity, and visual similarity—successfully comprise the RAG system with a significant performance drop.
Notably, punctuation insertion alone compromised the system in 86\% of the attacks, demonstrating \textit{GARAG}'s effectiveness in leveraging diverse perturbations for attacks.

\begin{table}[t!]
\centering
\small
\caption{\small Comparison with other search methods on NQ with Contriever and Llama-7b.}
\vspace{-.5em}
\centering
\resizebox{.9\columnwidth}{!}{
\begin{tabular}{l ccc c}
\toprule
 & \multicolumn{3}{c}{\textbf{ASR}} & \textbf{E2E}  \\ \cmidrule(l{2pt}r{2pt}){2-4} \cmidrule(l{2pt}r{2pt}){5-5} 
  & ASR$_R$ & ASR$_L$ & ASR$_T$ & EM \\ \midrule
\textbf{\textit{GARAG}} & 85.9 & 91.1 & 77.5 & 70.1  \\ 
\textbf{\textit{GARAG} on Retriever} & 96.6 & 18.0  & 18.0 & 94.4  \\ 
\textbf{\textit{GARAG} on LLM} & 33.2 & 100.0 & 33.2 & 85.2  \\ \midrule
\textbf{DS on Retriever} & 94.8 & 56.6 & 53.8 & 89.2 \\
\textbf{DS on LLM} & 16.0 & 100.0 & 16.0 & 90.4 \\ \midrule
\textbf{GS on Retriever} & 26.5 & 75.0 & 4.6 & 93.2 \\ 
\textbf{GS on LLM} & 4.9 & 96.2 & 17.8 & 97.2 \\ \bottomrule
\end{tabular}
}
\label{tab:5-5}
\vspace{-1.0em}
\end{table}

\paragraph{Comparison with Other Search Methods.}
We validated the effectiveness of our proposed method, \textit{GARAG}, by comparing it with two search methods based on word importance calculated through deletion scoring (DS) and gradient scoring (GS). Note that both methods can target only a single module. As shown in Table~\ref{tab:5-5}, these single-targeted methods fail to comprehensively search for adversarial documents across all modules. Even when implemented for single-module attacks, \textit{GARAG} achieves significantly higher ASR and lower E2E than other methods, demonstrating the genetic algorithm's effectiveness. This underscores the importance of attacking both retriever and reader rather than targeting a single module.

\begin{figure}[t]
\centering
\includegraphics[width=0.97\columnwidth]{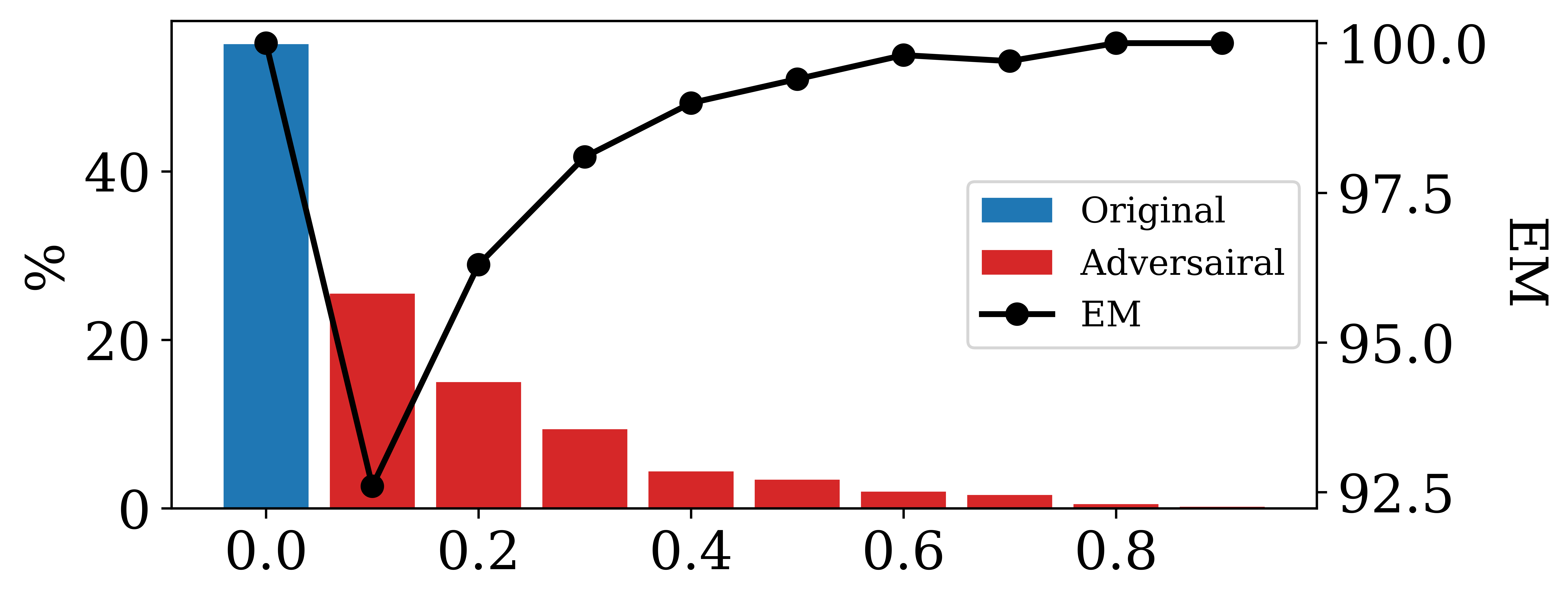}
\vspace{-.5em}
\caption{\small Distribution of grammatically correct documents among $\bm{d^*}$ on NQ with the Contriever and Llama2-7b.}
\label{fig:5-3}
\vspace{-1em}
\end{figure}

\begin{table*}[t!]
\caption{\small Case study with Contriever and Llama-7b, where perturbed texts are in\hlc[red!25]{red} and correct answers are in\hlc[blue!25]{blue}.}
\vspace{-.5em}
\small
\centering
\resizebox{\textwidth}{!}{
\renewcommand{\arraystretch}{0.8}
\begin{tabular}{cc}
\toprule 
\multicolumn{1}{p{.15\textwidth}}{\textbf{Question}} & \multicolumn{1}{p{.85\textwidth}}{Who sang the first line of `We Are The World'?} \\  \noalign{\vskip 0.5ex}\cdashline{1-2}\noalign{\vskip 0.5ex}
\multicolumn{1}{p{.15\textwidth}}{\textbf{Noisy Document}} & \multicolumn{1}{p{.85\textwidth}}{We Are the World lines in the sing's repetitive chorus proclaim, "We are the world, we are the children, we are the \hlc[red!25]{onss} who make a \hlc[red!25]{brighger} day, so let\'s start giving". "We Are the World" pens with \hlc[blue!25]{Lionel Richie}, \hlc[blue!25]{Stevie Wonder}, \hlc[blue!25]{Paul Simon}, \hlc[blue!25]{Kenny Rogers}, \hlc[blue!25]{James Ingram}, \hlc[blue!25]{Tina Turner}, and \hlc[blue!25]{Billy Joel} singing the first verse. Michael Jackson and Diana Ross \hlc[red!25]{f0llow}, completing the first \hlc[red!25]{choruc} together. Dionne Warwick, \hlc[red!25]{Willif} Nelson, and Al Jarreau \hlc[red!25]{singe} the second \hlc[red!25]{vers4}, before Bruce Springsteen, Kenny Loggins, Steve Perry, and Daryl Hall go through the second chorus.}  \\ \midrule
\multicolumn{1}{p{.15\textwidth}}{\textbf{Answer}} & \multicolumn{1}{p{.85\textwidth}}{Stevie Wonder, Tina Turner, Billy Joel, James Ingram, Kenny Rogers, Paul Simon, Lionel Richie} \\ \noalign{\vskip 0.5ex}\cdashline{1-2}\noalign{\vskip 0.5ex}
\multicolumn{1}{p{.15\textwidth}}{\textbf{Prediction}} & \multicolumn{1}{p{.85\textwidth}}{Michael Jackson} \\ 
\bottomrule

\end{tabular}
}
\label{tab:3}
\vspace{-1.5em}
\end{table*}



\paragraph{Defense Strategy.}
Various defense mechanisms against adversarial attacks in NLP have been proposed. Adversarial training, fine-tuning the model on adversarial samples, is a popular approach~\cite{adversarial_training}. However, this strategy is not practically viable for RAG systems, given the prohibitive training costs associated with models exceeding a billion parameters. Alternatively, a grammar checker is an effective defense against low-level perturbations within documents~\cite{punctuation}.
Our analysis, depicted in Figure~\ref{fig:5-3}, compares the grammatical correctness of original and adversarial documents via grammar checker model~\footnote{https://huggingface.co/imohammad12/GRS-Grammar-Checker-DeBerta} presented in~\citet{GRS}. 
It reveals that approximately 50\% of the original and clean samples are determined to be the nosiy documents containing grammatical errors. 
Also, even within the adversarial set, about 25\% of the samples maintain grammatical correctness at a low perturbation level.
This observation highlights a critical limitation: relying solely on a grammar checker would result in dismissing many original documents and accepting some adversarial ones.
Consequently, this underscores the limitations of grammar checkers as a standalone defense and points to more sophisticated and tailored defense strategies.

\paragraph{Case Study.}
We further qualitatively assess the impact of low-level textual perturbations within a document in Table~\ref{tab:3}. 
Note that since we ensure that the answer spans remain unperturbed, LLMs should ideally generate correct answers.
However, interestingly, an LLM fails to identify the correct answers, which are mentioned in the document, but instead generates an incorrect answer, ``Michael Jackson,'' included in the document. 
To this end, we would like to emphasize that addressing typographical errors is a complex challenge that requires many considerations in defense against the threat of typos, which seems relatively trivial. In our all experiments, we didn’t perturb the tokens included in the correct answer span, as shown in Table~\ref{tab:3}, and we empirically validated that RAG systems often can’t generate correct answers from the document, even including the correct answers. This poses a critical question: \textit{Should we discard such documents because of typographical errors or find ways to use this information effectively within them?} These considerations highlight the need for comprehensive and sophisticated defense strategies, underscoring the ongoing vulnerability within RAG systems.

In Appendix~\ref{sup:additional_results}, we provide detailed results of adversarial attacks for each dataset and analysis including comparing high-level perturbation attacks and attacking closed-source models.

\section{Conclusion}

In this work, we highlighted the importance of assessing the overall robustness of the retriever and reader components within the RAG system, particularly against noisy documents containing minor typos that are common in real-world databases.
Specifically, we proposed two objectives to evaluate the resilience of each component, focusing on their sequential dependencies. 
Furthermore, to simulate real-world noises with low-level perturbations, we introduced a novel adversarial attack method, \textit{GARAG}, incorporating a genetic algorithm. 
Our findings indicate that noisy documents critically hurt the RAG system, significantly degrading its performance. 
Although the retriever serves as a protective barrier for the reader, it still remains susceptible to minor disruptions. 
Our \textit{GARAG} shows promise as an adversarial attack strategy when assessing the holistic robustness of RAG systems against various low-level perturbations.

\section*{Acknowledgement}
This work was supported by Institute for Information and communications Technology Promotion (IITP) grant funded by the Korea government (No. 2018-0-00582, Prediction and augmentation of the credibility distribution via linguistic analysis and automated evidence document collection) and the Artificial intelligence industrial convergence cluster development project funded by the Ministry of Science and ICT (MSIT, Korea) \& Gwangju Metropolitan City.

\section*{Limitation}
In this work, we explored the robustness of the RAG system by using various recent open-source LLMs of different sizes, which are widely used as reader components in this system. However, due to our limited academic budget, we could not include much larger black-box LLMs such as the GPT series models, which have a hundred billion parameters. We believe that exploring the robustness of these LLMs as reader components would be a valuable line of future work.
Furthermore, \textit{GARAG} aims for the optimal adversarial document to be located within a holistic error zone, by simultaneously considering both retrieval and grounding errors. However, we would like to note that even though the adversarial document is located within the holistic error zone, this does not necessarily mean that the reader will always generate incorrect answers for every query, due to the auto-regressive nature of how reader models generate tokens. Nevertheless, as shown in the right figure of Figure~\ref{fig:3} and discussed in its analysis, we would like to emphasize that there is a clear correlation: a lower \(\mathcal{L}_{\textnormal{GPR}}\) value is associated with a higher likelihood of incorrect responses. 

\section*{Ethics Statement}
We designed a novel attack strategy for the purpose of building robust and safe RAG systems when deployed in the real world. However, given the potential for malicious users to exploit our \textit{GARAG} and deliberately attack the system, it is crucial to consider these scenarios. Therefore, to prevent such incidents, we also present a defense strategy, detailed in Figure~\ref{fig:5-3} and its analysis. Additionally, we believe that developing a range of defense strategies remains a critical area for future work.

\bibliography{custom}

\begin{thebibliography}{57}
\expandafter\ifx\csname natexlab\endcsname\relax\def\natexlab#1{#1}\fi

\bibitem[{Alzantot et~al.(2018)Alzantot, Sharma, Elgohary, Ho, Srivastava, and Chang}]{genetic_1}
Moustafa Alzantot, Yash Sharma, Ahmed Elgohary, Bo{-}Jhang Ho, Mani~B. Srivastava, and Kai{-}Wei Chang. 2018.
\newblock \href {https://doi.org/10.18653/V1/D18-1316} {Generating natural language adversarial examples}.
\newblock In \emph{Proceedings of the 2018 Conference on Empirical Methods in Natural Language Processing, Brussels, Belgium, October 31 - November 4, 2018}, pages 2890--2896. Association for Computational Linguistics.

\bibitem[{Asai et~al.(2024)Asai, Wu, Wang, Sil, and Hajishirzi}]{selfrag}
Akari Asai, Zeqiu Wu, Yizhong Wang, Avirup Sil, and Hannaneh Hajishirzi. 2024.
\newblock \href {https://openreview.net/forum?id=hSyW5go0v8} {Self-{RAG}: Learning to retrieve, generate, and critique through self-reflection}.
\newblock In \emph{The Twelfth International Conference on Learning Representations}.

\bibitem[{Baek et~al.(2023)Baek, Jeong, Kang, Park, and Hwang}]{KALMV}
Jinheon Baek, Soyeong Jeong, Minki Kang, Jong~C. Park, and Sung~Ju Hwang. 2023.
\newblock \href {https://aclanthology.org/2023.emnlp-main.107} {Knowledge-augmented language model verification}.
\newblock In \emph{Proceedings of the 2023 Conference on Empirical Methods in Natural Language Processing, {EMNLP} 2023, Singapore, December 6-10, 2023}, pages 1720--1736. Association for Computational Linguistics.

\bibitem[{Brown et~al.(2020)Brown, Mann, Ryder, Subbiah, Kaplan, Dhariwal, Neelakantan, Shyam, Sastry, Askell, Agarwal, Herbert{-}Voss, Krueger, Henighan, Child, Ramesh, Ziegler, Wu, Winter, Hesse, Chen, Sigler, Litwin, Gray, Chess, Clark, Berner, McCandlish, Radford, Sutskever, and Amodei}]{GPT3}
Tom~B. Brown, Benjamin Mann, Nick Ryder, Melanie Subbiah, Jared Kaplan, Prafulla Dhariwal, Arvind Neelakantan, Pranav Shyam, Girish Sastry, Amanda Askell, Sandhini Agarwal, Ariel Herbert{-}Voss, Gretchen Krueger, Tom Henighan, Rewon Child, Aditya Ramesh, Daniel~M. Ziegler, Jeffrey Wu, Clemens Winter, Christopher Hesse, Mark Chen, Eric Sigler, Mateusz Litwin, Scott Gray, Benjamin Chess, Jack Clark, Christopher Berner, Sam McCandlish, Alec Radford, Ilya Sutskever, and Dario Amodei. 2020.
\newblock \href {https://proceedings.neurips.cc/paper/2020/hash/1457c0d6bfcb4967418bfb8ac142f64a-Abstract.html} {Language models are few-shot learners}.
\newblock In \emph{Advances in Neural Information Processing Systems 33: Annual Conference on Neural Information Processing Systems 2020, NeurIPS 2020, December 6-12, 2020, virtual}.

\bibitem[{Chase(2022)}]{langchain}
Harrison Chase. 2022.
\newblock \href {https://github.com/langchain-ai/langchain} {{LangChain}}.

\bibitem[{Chen et~al.(2024)Chen, Lin, Han, and Sun}]{BenchmarkRAG}
Jiawei Chen, Hongyu Lin, Xianpei Han, and Le~Sun. 2024.
\newblock \href {https://doi.org/10.1609/AAAI.V38I16.29728} {Benchmarking large language models in retrieval-augmented generation}.
\newblock In \emph{Thirty-Eighth {AAAI} Conference on Artificial Intelligence, {AAAI} 2024, Thirty-Sixth Conference on Innovative Applications of Artificial Intelligence, {IAAI} 2024, Fourteenth Symposium on Educational Advances in Artificial Intelligence, {EAAI} 2014, February 20-27, 2024, Vancouver, Canada}, pages 17754--17762. {AAAI} Press.

\bibitem[{Chiang et~al.(2023)Chiang, Li, Lin, Sheng, Wu, Zhang, Zheng, Zhuang, Zhuang, Gonzalez, Stoica, and Xing}]{vicuna}
Wei-Lin Chiang, Zhuohan Li, Zi~Lin, Ying Sheng, Zhanghao Wu, Hao Zhang, Lianmin Zheng, Siyuan Zhuang, Yonghao Zhuang, Joseph~E. Gonzalez, Ion Stoica, and Eric~P. Xing. 2023.
\newblock \href {https://lmsys.org/blog/2023-03-30-vicuna/} {Vicuna: An open-source chatbot impressing gpt-4 with 90\%* chatgpt quality}.

\bibitem[{Cho et~al.(2023)Cho, Seo, Jeong, and Park}]{das}
Sukmin Cho, Jeongyeon Seo, Soyeong Jeong, and Jong~C. Park. 2023.
\newblock \href {https://aclanthology.org/2023.findings-emnlp.207} {Improving zero-shot reader by reducing distractions from irrelevant documents in open-domain question answering}.
\newblock In \emph{Findings of the Association for Computational Linguistics: {EMNLP} 2023, Singapore, December 6-10, 2023}, pages 3145--3157. Association for Computational Linguistics.

\bibitem[{Cuconasu et~al.(2024)Cuconasu, Trappolini, Siciliano, Filice, Campagnano, Maarek, Tonellotto, and Silvestri}]{PoN}
Florin Cuconasu, Giovanni Trappolini, Federico Siciliano, Simone Filice, Cesare Campagnano, Yoelle Maarek, Nicola Tonellotto, and Fabrizio Silvestri. 2024.
\newblock \href {https://doi.org/10.48550/ARXIV.2401.14887} {The power of noise: Redefining retrieval for {RAG} systems}.
\newblock \emph{arXiv preprint arXiv:2401.14887}, abs/2401.14887.

\bibitem[{Deb et~al.(2002)Deb, Agrawal, Pratap, and Meyarivan}]{NSGA}
Kalyanmoy Deb, Samir Agrawal, Amrit Pratap, and T.~Meyarivan. 2002.
\newblock \href {https://doi.org/10.1109/4235.996017} {A fast and elitist multiobjective genetic algorithm: {NSGA-II}}.
\newblock \emph{{IEEE} Trans. Evol. Comput.}, 6(2):182--197.

\bibitem[{Dehghan et~al.(2022)Dehghan, Kumar, and Golab}]{GRS}
Mohammad Dehghan, Dhruv Kumar, and Lukasz Golab. 2022.
\newblock \href {https://doi.org/10.18653/V1/2022.FINDINGS-ACL.77} {{GRS:} combining generation and revision in unsupervised sentence simplification}.
\newblock In \emph{Findings of the Association for Computational Linguistics: {ACL} 2022, Dublin, Ireland, May 22-27, 2022}, pages 949--960. Association for Computational Linguistics.

\bibitem[{Ebrahimi et~al.(2018)Ebrahimi, Rao, Lowd, and Dou}]{gradient_2}
Javid Ebrahimi, Anyi Rao, Daniel Lowd, and Dejing Dou. 2018.
\newblock \href {https://doi.org/10.18653/V1/P18-2006} {Hotflip: White-box adversarial examples for text classification}.
\newblock In \emph{Proceedings of the 56th Annual Meeting of the Association for Computational Linguistics, {ACL} 2018, Melbourne, Australia, July 15-20, 2018, Volume 2: Short Papers}, pages 31--36. Association for Computational Linguistics.

\bibitem[{Eger and Benz(2020)}]{zeroe}
Steffen Eger and Yannik Benz. 2020.
\newblock \href {https://aclanthology.org/2020.aacl-main.79/} {From hero to z{\'{e}}roe: {A} benchmark of low-level adversarial attacks}.
\newblock In \emph{Proceedings of the 1st Conference of the Asia-Pacific Chapter of the Association for Computational Linguistics and the 10th International Joint Conference on Natural Language Processing, {AACL/IJCNLP} 2020, Suzhou, China, December 4-7, 2020}, pages 786--803. Association for Computational Linguistics.

\bibitem[{Eger et~al.(2019)Eger, Sahin, R{\"{u}}ckl{\'{e}}, Lee, Schulz, Mesgar, Swarnkar, Simpson, and Gurevych}]{viper}
Steffen Eger, G{\"{o}}zde~G{\"{u}}l Sahin, Andreas R{\"{u}}ckl{\'{e}}, Ji{-}Ung Lee, Claudia Schulz, Mohsen Mesgar, Krishnkant Swarnkar, Edwin Simpson, and Iryna Gurevych. 2019.
\newblock \href {https://doi.org/10.18653/V1/N19-1165} {Text processing like humans do: Visually attacking and shielding {NLP} systems}.
\newblock In \emph{Proceedings of the 2019 Conference of the North American Chapter of the Association for Computational Linguistics: Human Language Technologies, {NAACL-HLT} 2019, Minneapolis, MN, USA, June 2-7, 2019, Volume 1 (Long and Short Papers)}, pages 1634--1647. Association for Computational Linguistics.

\bibitem[{Faruqui et~al.(2018)Faruqui, Pavlick, Tenney, and Das}]{WikipediaTypo}
Manaal Faruqui, Ellie Pavlick, Ian Tenney, and Dipanjan Das. 2018.
\newblock \href {https://doi.org/10.18653/V1/D18-1028} {Wikiatomicedits: {A} multilingual corpus of wikipedia edits for modeling language and discourse}.
\newblock In \emph{Proceedings of the 2018 Conference on Empirical Methods in Natural Language Processing, Brussels, Belgium, October 31 - November 4, 2018}, pages 305--315. Association for Computational Linguistics.

\bibitem[{Formento et~al.(2023)Formento, Foo, Luu, and Ng}]{punctuation}
Brian Formento, Chuan{-}Sheng Foo, Anh~Tuan Luu, and See{-}Kiong Ng. 2023.
\newblock \href {https://doi.org/10.18653/V1/2023.FINDINGS-EACL.1} {Using punctuation as an adversarial attack on deep learning-based {NLP} systems: An empirical study}.
\newblock In \emph{Findings of the Association for Computational Linguistics: {EACL} 2023, Dubrovnik, Croatia, May 2-6, 2023}, pages 1--34. Association for Computational Linguistics.

\bibitem[{Gao et~al.(2018)Gao, Lanchantin, Soffa, and Qi}]{deletion_search}
Ji~Gao, Jack Lanchantin, Mary~Lou Soffa, and Yanjun Qi. 2018.
\newblock \href {https://doi.org/10.1109/SPW.2018.00016} {Black-box generation of adversarial text sequences to evade deep learning classifiers}.
\newblock In \emph{2018 {IEEE} Security and Privacy Workshops, {SP} Workshops 2018, San Francisco, CA, USA, May 24, 2018}, pages 50--56. {IEEE} Computer Society.

\bibitem[{Izacard et~al.(2022)Izacard, Caron, Hosseini, Riedel, Bojanowski, Joulin, and Grave}]{contriever}
Gautier Izacard, Mathilde Caron, Lucas Hosseini, Sebastian Riedel, Piotr Bojanowski, Armand Joulin, and Edouard Grave. 2022.
\newblock \href {https://openreview.net/forum?id=jKN1pXi7b0} {Unsupervised dense information retrieval with contrastive learning}.
\newblock \emph{Trans. Mach. Learn. Res.}, 2022.

\bibitem[{Jeong et~al.(2024)Jeong, Baek, Cho, Hwang, and Park}]{adaptiverag}
Soyeong Jeong, Jinheon Baek, Sukmin Cho, Sung~Ju Hwang, and Jong~C. Park. 2024.
\newblock \href {https://openreview.net/forum?id=RYyLwb4NcY} {Adaptive-{RAG}: Learning to adapt retrieval-augmented large language models through question complexity}.
\newblock In \emph{2024 Annual Conference of the North American Chapter of the Association for Computational Linguistics}.

\bibitem[{Jiang et~al.(2023)Jiang, Sablayrolles, Mensch, Bamford, Chaplot, de~Las~Casas, Bressand, Lengyel, Lample, Saulnier, Lavaud, Lachaux, Stock, Scao, Lavril, Wang, Lacroix, and Sayed}]{mistral}
Albert~Q. Jiang, Alexandre Sablayrolles, Arthur Mensch, Chris Bamford, Devendra~Singh Chaplot, Diego de~Las~Casas, Florian Bressand, Gianna Lengyel, Guillaume Lample, Lucile Saulnier, L{\'{e}}lio~Renard Lavaud, Marie{-}Anne Lachaux, Pierre Stock, Teven~Le Scao, Thibaut Lavril, Thomas Wang, Timoth{\'{e}}e Lacroix, and William~El Sayed. 2023.
\newblock \href {https://doi.org/10.48550/ARXIV.2310.06825} {Mistral 7b}.
\newblock \emph{arXiv preprint arXiv:2310.06825}, abs/2310.06825.

\bibitem[{Jin et~al.(2020)Jin, Jin, Zhou, and Szolovits}]{textfooler}
Di~Jin, Zhijing Jin, Joey~Tianyi Zhou, and Peter Szolovits. 2020.
\newblock \href {https://doi.org/10.1609/AAAI.V34I05.6311} {Is {BERT} really robust? {A} strong baseline for natural language attack on text classification and entailment}.
\newblock In \emph{The Thirty-Fourth {AAAI} Conference on Artificial Intelligence, {AAAI} 2020, The Thirty-Second Innovative Applications of Artificial Intelligence Conference, {IAAI} 2020, The Tenth {AAAI} Symposium on Educational Advances in Artificial Intelligence, {EAAI} 2020, New York, NY, USA, February 7-12, 2020}, pages 8018--8025. {AAAI} Press.

\bibitem[{Joshi et~al.(2017)Joshi, Choi, Weld, and Zettlemoyer}]{TQA}
Mandar Joshi, Eunsol Choi, Daniel~S. Weld, and Luke Zettlemoyer. 2017.
\newblock \href {https://doi.org/10.18653/V1/P17-1147} {Triviaqa: {A} large scale distantly supervised challenge dataset for reading comprehension}.
\newblock In \emph{Proceedings of the 55th Annual Meeting of the Association for Computational Linguistics, {ACL} 2017, Vancouver, Canada, July 30 - August 4, Volume 1: Long Papers}, pages 1601--1611. Association for Computational Linguistics.

\bibitem[{Karpukhin et~al.(2020)Karpukhin, Oguz, Min, Lewis, Wu, Edunov, Chen, and Yih}]{DPR}
Vladimir Karpukhin, Barlas Oguz, Sewon Min, Patrick S.~H. Lewis, Ledell Wu, Sergey Edunov, Danqi Chen, and Wen{-}tau Yih. 2020.
\newblock \href {https://doi.org/10.18653/V1/2020.EMNLP-MAIN.550} {Dense passage retrieval for open-domain question answering}.
\newblock In \emph{Proceedings of the 2020 Conference on Empirical Methods in Natural Language Processing, {EMNLP} 2020, Online, November 16-20, 2020}, pages 6769--6781. Association for Computational Linguistics.

\bibitem[{Kasai et~al.(2023)Kasai, Sakaguchi, Takahashi, Bras, Asai, Yu, Radev, Smith, Choi, and Inui}]{realtime}
Jungo Kasai, Keisuke Sakaguchi, Yoichi Takahashi, Ronan~Le Bras, Akari Asai, Xinyan Yu, Dragomir Radev, Noah~A. Smith, Yejin Choi, and Kentaro Inui. 2023.
\newblock \href {http://papers.nips.cc/paper\_files/paper/2023/hash/9941624ef7f867a502732b5154d30cb7-Abstract-Datasets\_and\_Benchmarks.html} {Realtime {QA:} what's the answer right now?}
\newblock In \emph{Advances in Neural Information Processing Systems 36: Annual Conference on Neural Information Processing Systems 2023, NeurIPS 2023, New Orleans, LA, USA, December 10 - 16, 2023}.

\bibitem[{Kwiatkowski et~al.(2019)Kwiatkowski, Palomaki, Redfield, Collins, Parikh, Alberti, Epstein, Polosukhin, Devlin, Lee, Toutanova, Jones, Kelcey, Chang, Dai, Uszkoreit, Le, and Petrov}]{NQ}
Tom Kwiatkowski, Jennimaria Palomaki, Olivia Redfield, Michael Collins, Ankur~P. Parikh, Chris Alberti, Danielle Epstein, Illia Polosukhin, Jacob Devlin, Kenton Lee, Kristina Toutanova, Llion Jones, Matthew Kelcey, Ming{-}Wei Chang, Andrew~M. Dai, Jakob Uszkoreit, Quoc Le, and Slav Petrov. 2019.
\newblock \href {https://doi.org/10.1162/TACL\_A\_00276} {Natural questions: a benchmark for question answering research}.
\newblock \emph{Trans. Assoc. Comput. Linguistics}, 7:452--466.

\bibitem[{Lazaridou et~al.(2022)Lazaridou, Gribovskaya, Stokowiec, and Grigorev}]{internetaug}
Angeliki Lazaridou, Elena Gribovskaya, Wojciech Stokowiec, and Nikolai Grigorev. 2022.
\newblock \href {https://doi.org/10.48550/ARXIV.2203.05115} {Internet-augmented language models through few-shot prompting for open-domain question answering}.
\newblock \emph{arXiv preprint arXiv:2203.05115}, abs/2203.05115.

\bibitem[{Le et~al.(2022)Le, Lee, Yen, Hu, and Lee}]{anthro}
Thai Le, Jooyoung Lee, Kevin Yen, Yifan Hu, and Dongwon Lee. 2022.
\newblock \href {https://doi.org/10.18653/V1/2022.FINDINGS-ACL.232} {Perturbations in the wild: Leveraging human-written text perturbations for realistic adversarial attack and defense}.
\newblock In \emph{Findings of the Association for Computational Linguistics: {ACL} 2022, Dublin, Ireland, May 22-27, 2022}, pages 2953--2965. Association for Computational Linguistics.

\bibitem[{Le et~al.(2023)Le, Ye, Hu, and Lee}]{CryptText}
Thai Le, Yiran Ye, Yifan Hu, and Dongwon Lee. 2023.
\newblock \href {https://doi.org/10.1109/ICDE55515.2023.00287} {Cryptext: Database and interactive toolkit of human-written text perturbations in the wild}.
\newblock In \emph{39th {IEEE} International Conference on Data Engineering, {ICDE} 2023, Anaheim, CA, USA, April 3-7, 2023}, pages 3639--3642. {IEEE}.

\bibitem[{Lee et~al.(2023)Lee, Joo, Kim, Jang, Kim, On, and Seo}]{ground}
Hyunji Lee, Se~June Joo, Chaeeun Kim, Joel Jang, Doyoung Kim, Kyoung{-}Woon On, and Minjoon Seo. 2023.
\newblock \href {https://doi.org/10.48550/ARXIV.2311.09069} {How well do large language models truly ground?}
\newblock \emph{arXiv preprint arXiv:2311.09069}, abs/2311.09069.

\bibitem[{Lewis et~al.(2020)Lewis, Perez, Piktus, Petroni, Karpukhin, Goyal, K{\"{u}}ttler, Lewis, Yih, Rockt{\"{a}}schel, Riedel, and Kiela}]{rag}
Patrick S.~H. Lewis, Ethan Perez, Aleksandra Piktus, Fabio Petroni, Vladimir Karpukhin, Naman Goyal, Heinrich K{\"{u}}ttler, Mike Lewis, Wen{-}tau Yih, Tim Rockt{\"{a}}schel, Sebastian Riedel, and Douwe Kiela. 2020.
\newblock \href {https://proceedings.neurips.cc/paper/2020/hash/6b493230205f780e1bc26945df7481e5-Abstract.html} {Retrieval-augmented generation for knowledge-intensive {NLP} tasks}.
\newblock In \emph{Advances in Neural Information Processing Systems 33: Annual Conference on Neural Information Processing Systems 2020, NeurIPS 2020, December 6-12, 2020, virtual}.

\bibitem[{Li et~al.(2023{\natexlab{a}})Li, Cheng, Zhao, Nie, and Wen}]{HaluEval}
Junyi Li, Xiaoxue Cheng, Xin Zhao, Jian{-}Yun Nie, and Ji{-}Rong Wen. 2023{\natexlab{a}}.
\newblock \href {https://aclanthology.org/2023.emnlp-main.397} {Halueval: {A} large-scale hallucination evaluation benchmark for large language models}.
\newblock In \emph{Proceedings of the 2023 Conference on Empirical Methods in Natural Language Processing, {EMNLP} 2023, Singapore, December 6-10, 2023}, pages 6449--6464. Association for Computational Linguistics.

\bibitem[{Li et~al.(2023{\natexlab{b}})Li, Liu, Gao, and Buntine}]{ood}
Xinzhe Li, Ming Liu, Shang Gao, and Wray~L. Buntine. 2023{\natexlab{b}}.
\newblock \href {https://doi.org/10.24963/IJCAI.2023/749} {A survey on out-of-distribution evaluation of neural {NLP} models}.
\newblock In \emph{Proceedings of the Thirty-Second International Joint Conference on Artificial Intelligence, {IJCAI} 2023, 19th-25th August 2023, Macao, SAR, China}, pages 6683--6691. ijcai.org.

\bibitem[{Liu(2022)}]{Llamaindex}
Jerry Liu. 2022.
\newblock \href {https://github.com/jerryjliu/llama_index} {{LlamaIndex}}.

\bibitem[{Long et~al.(2024)Long, Deng, Gan, Wang, and Pan}]{backdoor}
Quanyu Long, Yue Deng, Leilei Gan, Wenya Wang, and Sinno~Jialin Pan. 2024.
\newblock \href {https://doi.org/10.48550/ARXIV.2402.13532} {Backdoor attacks on dense passage retrievers for disseminating misinformation}.
\newblock \emph{arXiv preprint arXiv:2402.13532}, abs/2402.13532.

\bibitem[{Mallen et~al.(2023)Mallen, Asai, Zhong, Das, Khashabi, and Hajishirzi}]{popqa}
Alex Mallen, Akari Asai, Victor Zhong, Rajarshi Das, Daniel Khashabi, and Hannaneh Hajishirzi. 2023.
\newblock \href {https://doi.org/10.18653/V1/2023.ACL-LONG.546} {When not to trust language models: Investigating effectiveness of parametric and non-parametric memories}.
\newblock In \emph{Proceedings of the 61st Annual Meeting of the Association for Computational Linguistics (Volume 1: Long Papers), {ACL} 2023, Toronto, Canada, July 9-14, 2023}, pages 9802--9822. Association for Computational Linguistics.

\bibitem[{OpenAI(2023{\natexlab{a}})}]{Plugin}
OpenAI. 2023{\natexlab{a}}.
\newblock \href {https://openai.com/blog/chatgpt-plugins} {Chatgpt plugins}.

\bibitem[{OpenAI(2023{\natexlab{b}})}]{GPT4}
OpenAI. 2023{\natexlab{b}}.
\newblock \href {https://doi.org/10.48550/ARXIV.2303.08774} {{GPT-4} technical report}.
\newblock \emph{arXiv preprint arXiv:2303.08774}, abs/2303.08774.

\bibitem[{Piktus et~al.(2021)Piktus, Petroni, Karpukhin, Okhonko, Broscheit, Izacard, Lewis, Oguz, Grave, Yih, and Riedel}]{Web_Oyster}
Aleksandra Piktus, Fabio Petroni, Vladimir Karpukhin, Dmytro Okhonko, Samuel Broscheit, Gautier Izacard, Patrick S.~H. Lewis, Barlas Oguz, Edouard Grave, Wen{-}tau Yih, and Sebastian Riedel. 2021.
\newblock \href {http://arxiv.org/abs/2112.09924} {The web is your oyster - knowledge-intensive {NLP} against a very large web corpus}.
\newblock \emph{arXiv preprint arXiv:2112.09924}, abs/2112.09924.

\bibitem[{Press et~al.(2023)Press, Zhang, Min, Schmidt, Smith, and Lewis}]{self-ask}
Ofir Press, Muru Zhang, Sewon Min, Ludwig Schmidt, Noah~A. Smith, and Mike Lewis. 2023.
\newblock \href {https://aclanthology.org/2023.findings-emnlp.378} {Measuring and narrowing the compositionality gap in language models}.
\newblock In \emph{Findings of the Association for Computational Linguistics: {EMNLP} 2023, Singapore, December 6-10, 2023}, pages 5687--5711. Association for Computational Linguistics.

\bibitem[{Qin et~al.(2024)Qin, Liang, Ye, Zhu, Yan, Lu, Lin, Cong, Tang, Qian, Zhao, Hong, Tian, Xie, Zhou, Gerstein, dahai li, Liu, and Sun}]{ToolLLM}
Yujia Qin, Shihao Liang, Yining Ye, Kunlun Zhu, Lan Yan, Yaxi Lu, Yankai Lin, Xin Cong, Xiangru Tang, Bill Qian, Sihan Zhao, Lauren Hong, Runchu Tian, Ruobing Xie, Jie Zhou, Mark Gerstein, dahai li, Zhiyuan Liu, and Maosong Sun. 2024.
\newblock \href {https://openreview.net/forum?id=dHng2O0Jjr} {Tool{LLM}: Facilitating large language models to master 16000+ real-world {API}s}.
\newblock In \emph{The Twelfth International Conference on Learning Representations}.

\bibitem[{Rajpurkar et~al.(2016)Rajpurkar, Zhang, Lopyrev, and Liang}]{SQD}
Pranav Rajpurkar, Jian Zhang, Konstantin Lopyrev, and Percy Liang. 2016.
\newblock \href {https://doi.org/10.18653/V1/D16-1264} {Squad: 100, 000+ questions for machine comprehension of text}.
\newblock In \emph{Proceedings of the 2016 Conference on Empirical Methods in Natural Language Processing, {EMNLP} 2016, Austin, Texas, USA, November 1-4, 2016}, pages 2383--2392. The Association for Computational Linguistics.

\bibitem[{Ribeiro et~al.(2018)Ribeiro, Singh, and Guestrin}]{paraphrase}
Marco~T{\'{u}}lio Ribeiro, Sameer Singh, and Carlos Guestrin. 2018.
\newblock \href {https://doi.org/10.18653/V1/P18-1079} {Semantically equivalent adversarial rules for debugging {NLP} models}.
\newblock In \emph{Proceedings of the 56th Annual Meeting of the Association for Computational Linguistics, {ACL} 2018, Melbourne, Australia, July 15-20, 2018, Volume 1: Long Papers}, pages 856--865. Association for Computational Linguistics.

\bibitem[{Thakur et~al.(2023)Thakur, Bonifacio, Zhang, Ogundepo, Kamalloo, Alfonso{-}Hermelo, Li, Liu, Chen, Rezagholizadeh, and Lin}]{nomiracl}
Nandan Thakur, Luiz Bonifacio, Xinyu Zhang, Odunayo Ogundepo, Ehsan Kamalloo, David Alfonso{-}Hermelo, Xiaoguang Li, Qun Liu, Boxing Chen, Mehdi Rezagholizadeh, and Jimmy Lin. 2023.
\newblock \href {https://doi.org/10.48550/ARXIV.2312.11361} {Nomiracl: Knowing when you don't know for robust multilingual retrieval-augmented generation}.
\newblock \emph{arXiv preprint arXiv:2312.11361}, abs/2312.11361.

\bibitem[{Thakur et~al.(2021)Thakur, Reimers, R{\"{u}}ckl{\'{e}}, Srivastava, and Gurevych}]{BEIR}
Nandan Thakur, Nils Reimers, Andreas R{\"{u}}ckl{\'{e}}, Abhishek Srivastava, and Iryna Gurevych. 2021.
\newblock \href {https://datasets-benchmarks-proceedings.neurips.cc/paper/2021/hash/65b9eea6e1cc6bb9f0cd2a47751a186f-Abstract-round2.html} {{BEIR:} {A} heterogeneous benchmark for zero-shot evaluation of information retrieval models}.
\newblock In \emph{Proceedings of the Neural Information Processing Systems Track on Datasets and Benchmarks 1, NeurIPS Datasets and Benchmarks 2021, December 2021, virtual}.

\bibitem[{Touvron et~al.(2023)Touvron, Martin, Stone, Albert, Almahairi, Babaei, Bashlykov, Batra, Bhargava, Bhosale, Bikel, Blecher, Canton{-}Ferrer, Chen, Cucurull, Esiobu, Fernandes, Fu, Fu, Fuller, Gao, Goswami, Goyal, Hartshorn, Hosseini, Hou, Inan, Kardas, Kerkez, Khabsa, Kloumann, Korenev, Koura, Lachaux, Lavril, Lee, Liskovich, Lu, Mao, Martinet, Mihaylov, Mishra, Molybog, Nie, Poulton, Reizenstein, Rungta, Saladi, Schelten, Silva, Smith, Subramanian, Tan, Tang, Taylor, Williams, Kuan, Xu, Yan, Zarov, Zhang, Fan, Kambadur, Narang, Rodriguez, Stojnic, Edunov, and Scialom}]{Llama2}
Hugo Touvron, Louis Martin, Kevin Stone, Peter Albert, Amjad Almahairi, Yasmine Babaei, Nikolay Bashlykov, Soumya Batra, Prajjwal Bhargava, Shruti Bhosale, Dan Bikel, Lukas Blecher, Cristian Canton{-}Ferrer, Moya Chen, Guillem Cucurull, David Esiobu, Jude Fernandes, Jeremy Fu, Wenyin Fu, Brian Fuller, Cynthia Gao, Vedanuj Goswami, Naman Goyal, Anthony Hartshorn, Saghar Hosseini, Rui Hou, Hakan Inan, Marcin Kardas, Viktor Kerkez, Madian Khabsa, Isabel Kloumann, Artem Korenev, Punit~Singh Koura, Marie{-}Anne Lachaux, Thibaut Lavril, Jenya Lee, Diana Liskovich, Yinghai Lu, Yuning Mao, Xavier Martinet, Todor Mihaylov, Pushkar Mishra, Igor Molybog, Yixin Nie, Andrew Poulton, Jeremy Reizenstein, Rashi Rungta, Kalyan Saladi, Alan Schelten, Ruan Silva, Eric~Michael Smith, Ranjan Subramanian, Xiaoqing~Ellen Tan, Binh Tang, Ross Taylor, Adina Williams, Jian~Xiang Kuan, Puxin Xu, Zheng Yan, Iliyan Zarov, Yuchen Zhang, Angela Fan, Melanie Kambadur, Sharan Narang, Aur{\'{e}}lien Rodriguez, Robert Stojnic, Sergey Edunov,
  and Thomas Scialom. 2023.
\newblock \href {https://doi.org/10.48550/ARXIV.2307.09288} {Llama 2: Open foundation and fine-tuned chat models}.
\newblock \emph{arXiv preprint arXiv:2307.09288}, abs/2307.09288.

\bibitem[{Wang et~al.(2023)Wang, Chen, Pei, Xie, Kang, Zhang, Xu, Xiong, Dutta, Schaeffer, Truong, Arora, Mazeika, Hendrycks, Lin, Cheng, Koyejo, Song, and Li}]{decodtrust}
Boxin Wang, Weixin Chen, Hengzhi Pei, Chulin Xie, Mintong Kang, Chenhui Zhang, Chejian Xu, Zidi Xiong, Ritik Dutta, Rylan Schaeffer, Sang~T. Truong, Simran Arora, Mantas Mazeika, Dan Hendrycks, Zinan Lin, Yu~Cheng, Sanmi Koyejo, Dawn Song, and Bo~Li. 2023.
\newblock \href {http://papers.nips.cc/paper\_files/paper/2023/hash/63cb9921eecf51bfad27a99b2c53dd6d-Abstract-Datasets\_and\_Benchmarks.html} {Decodingtrust: {A} comprehensive assessment of trustworthiness in {GPT} models}.
\newblock In \emph{Advances in Neural Information Processing Systems 36: Annual Conference on Neural Information Processing Systems 2023, NeurIPS 2023, New Orleans, LA, USA, December 10 - 16, 2023}.

\bibitem[{Wang et~al.(2021)Wang, Xu, Wang, Gan, Cheng, Gao, Awadallah, and Li}]{advGlue}
Boxin Wang, Chejian Xu, Shuohang Wang, Zhe Gan, Yu~Cheng, Jianfeng Gao, Ahmed~Hassan Awadallah, and Bo~Li. 2021.
\newblock \href {https://datasets-benchmarks-proceedings.neurips.cc/paper/2021/hash/335f5352088d7d9bf74191e006d8e24c-Abstract-round2.html} {Adversarial {GLUE:} {A} multi-task benchmark for robustness evaluation of language models}.
\newblock In \emph{Proceedings of the Neural Information Processing Systems Track on Datasets and Benchmarks 1, NeurIPS Datasets and Benchmarks 2021, December 2021, virtual}.

\bibitem[{Wang et~al.(2024)Wang, Ren, Li, Zhao, Liu, and Wen}]{REAR}
Yuhao Wang, Ruiyang Ren, Junyi Li, Wayne~Xin Zhao, Jing Liu, and Ji{-}Rong Wen. 2024.
\newblock \href {https://doi.org/10.48550/ARXIV.2402.17497} {{REAR:} {A} relevance-aware retrieval-augmented framework for open-domain question answering}.
\newblock \emph{arXiv preprint arXiv:2402.17497}, abs/2402.17497.

\bibitem[{Williams and Li(2023)}]{genetic_3}
Phoenix~Neale Williams and Ke~Li. 2023.
\newblock \href {https://doi.org/10.1109/CVPR52729.2023.01183} {Black-box sparse adversarial attack via multi-objective optimisation {CVPR} proceedings}.
\newblock In \emph{{IEEE/CVF} Conference on Computer Vision and Pattern Recognition, {CVPR} 2023, Vancouver, BC, Canada, June 17-24, 2023}, pages 12291--12301. {IEEE}.

\bibitem[{Yoo and Qi(2021{\natexlab{a}})}]{gradient_search}
Jin~Yong Yoo and Yanjun Qi. 2021{\natexlab{a}}.
\newblock \href {https://doi.org/10.18653/V1/2021.FINDINGS-EMNLP.81} {Towards improving adversarial training of {NLP} models}.
\newblock In \emph{Findings of the Association for Computational Linguistics: {EMNLP} 2021, Virtual Event / Punta Cana, Dominican Republic, 16-20 November, 2021}, pages 945--956. Association for Computational Linguistics.

\bibitem[{Yoo and Qi(2021{\natexlab{b}})}]{adversarial_training}
Jin~Yong Yoo and Yanjun Qi. 2021{\natexlab{b}}.
\newblock \href {https://doi.org/10.18653/V1/2021.FINDINGS-EMNLP.81} {Towards improving adversarial training of {NLP} models}.
\newblock In \emph{Findings of the Association for Computational Linguistics: {EMNLP} 2021, Virtual Event / Punta Cana, Dominican Republic, 16-20 November, 2021}, pages 945--956. Association for Computational Linguistics.

\bibitem[{Yoran et~al.(2024)Yoran, Wolfson, Ram, and Berant}]{JudgeThenGen}
Ori Yoran, Tomer Wolfson, Ori Ram, and Jonathan Berant. 2024.
\newblock \href {https://openreview.net/forum?id=ZS4m74kZpH} {Making retrieval-augmented language models robust to irrelevant context}.
\newblock In \emph{The Twelfth International Conference on Learning Representations}.

\bibitem[{Zang et~al.(2020)Zang, Qi, Yang, Liu, Zhang, Liu, and Sun}]{genetic_2}
Yuan Zang, Fanchao Qi, Chenghao Yang, Zhiyuan Liu, Meng Zhang, Qun Liu, and Maosong Sun. 2020.
\newblock \href {https://doi.org/10.18653/V1/2020.ACL-MAIN.540} {Word-level textual adversarial attacking as combinatorial optimization}.
\newblock In \emph{Proceedings of the 58th Annual Meeting of the Association for Computational Linguistics, {ACL} 2020, Online, July 5-10, 2020}, pages 6066--6080. Association for Computational Linguistics.

\bibitem[{Zhang et~al.(2020)Zhang, Sheng, Alhazmi, and Li}]{adversarial_survey}
Wei~Emma Zhang, Quan~Z. Sheng, Ahoud Alhazmi, and Chenliang Li. 2020.
\newblock \href {https://doi.org/10.1145/3374217} {Adversarial attacks on deep-learning models in natural language processing: {A} survey}.
\newblock \emph{{ACM} Trans. Intell. Syst. Technol.}, 11(3):24:1--24:41.

\bibitem[{Zhong et~al.(2023)Zhong, Huang, Wettig, and Chen}]{PoisonRetriever}
Zexuan Zhong, Ziqing Huang, Alexander Wettig, and Danqi Chen. 2023.
\newblock \href {https://aclanthology.org/2023.emnlp-main.849} {Poisoning retrieval corpora by injecting adversarial passages}.
\newblock In \emph{Proceedings of the 2023 Conference on Empirical Methods in Natural Language Processing, {EMNLP} 2023, Singapore, December 6-10, 2023}, pages 13764--13775. Association for Computational Linguistics.

\bibitem[{Zhu et~al.(2023)Zhu, Wang, Zhou, Wang, Chen, Wang, Yang, Ye, Gong, Zhang, and Xie}]{Promptbench}
Kaijie Zhu, Jindong Wang, Jiaheng Zhou, Zichen Wang, Hao Chen, Yidong Wang, Linyi Yang, Wei Ye, Neil~Zhenqiang Gong, Yue Zhang, and Xing Xie. 2023.
\newblock \href {https://doi.org/10.48550/ARXIV.2306.04528} {Promptbench: Towards evaluating the robustness of large language models on adversarial prompts}.
\newblock \emph{arXiv preprint arXiv:2306.04528}, abs/2306.04528.

\bibitem[{Zou et~al.(2024)Zou, Geng, Wang, and Jia}]{PoisonRAG}
Wei Zou, Runpeng Geng, Binghui Wang, and Jinyuan Jia. 2024.
\newblock \href {https://doi.org/10.48550/ARXIV.2402.07867} {Poisonedrag: Knowledge poisoning attacks to retrieval-augmented generation of large language models}.
\newblock \emph{arXiv preprint arXiv:2402.07867}, abs/2402.07867.

\end{thebibliography}

\appendix

\clearpage

\section{Implementation Detail}\label{sup:operation}

\subsection{Operations}

We explore four types of low-level perturbations, capturing the unpredictable and diverse nature of textual typos from~\citet{zeroe}. The operations of transformation function $f$ in our work are as follows:
\begin{itemize}
    \item \textbf{Inner-Shuffle}: Randomly shuffles the letters within a subsequence of a word token, limited to words with more than three characters.
    \item \textbf{Truncate}: Removes a random number of letters from a word token's beginning or end. This operation is restricted to words with more than three characters, with a maximum of three characters removed.
    \item \textbf{Keyboard Typo}: Substitutes a letter with its adjacent counterpart on an English keyboard layout to simulate human typing errors. Only one character per word is replaced.
    \item \textbf{Natural Typo}: Replaces letters based on common human errors derived from Wikipedia's edit history. This operation encompasses a variety of error types, including phonetic errors, omissions, morphological errors, and their combinations.
\end{itemize}
 
Additionally, we explore other types of low-level perturbations, such as punctuation insertion and phonetic and visual similarity. The operations of these low-level perturbations are as follows:
\begin{itemize}
    \item \textbf{Punctuation Insertion}: Insert random punctuations into the beginning or end of a word token. We insert a maximum of three identical punctuations into the beginning or end of the word. Exploited punctuations are " ,.'!?; ".
    \item \textbf{Phonetic Similarity}: Swap the characters in a word into the other tokens having phonetic similarity with the original ones. We exploit two types of phonetic similarity attacks from~\citet{zeroe} and~\citet{anthro}.
    \item \textbf{Visual Similarity}: Swap the characters in a word into the other tokens having visual similarity with the original ones. We exploit two types of phonetic similarity attacks from~\citet{viper}.
\end{itemize}

\subsection{Details of Attack Objectives}

In this section, we explain the details of the attack objectives: the Relevance Score Ratio (RSR) and the Generation Probability Ratio (GPR).

First, the Relevance Score Ratio (RSR) calculates the ratio of the relevance score from the adversarial document \(\bm{d'}\) to the score from the original document \(\bm{d}\) for a given query \(\bm{q}\). This ratio measures the superiority of the relevance score for \(\bm{q}\) between \(\bm{d}\) and \(\bm{d'}\). For instance, if the RSR value is below 1, the relevance score from \(\bm{d'}\) is higher than that from \(\bm{d}\). Although this ratio is relative to the original document \(\bm{d}\) and does not capture the actual rank in the retriever corpus, we validated the actual performance degradation of the retriever models, as shown in Table~\ref{tab:2}. 

The Generation Probability Ratio (GPR) calculates the ratio of the generation probabilities of the correct answer \(\bm{a}\) from the original pair \((\bm{d},\bm{q})\) to the probability from the adversarial pair \((\bm{d'},\bm{q})\). The generation probability of the answer \(\bm{a}\) for a document-query pair \((\bm{d},\bm{q})\) is the joint probability over the answer tokens in \(\bm{a}\), represented as \(p(\bm{a}|\bm{d},\bm{q}) = \prod_{i=1}^L p(a_{i}|a_{<i},\bm{d},\bm{q})\).
This ratio measures the likelihood that the adversarial document will cause the LLM to generate the correct answer \(\bm{a}\) compared to the original document \(\bm{d}\). For instance, if the GPR value is below 1, the adversarial document \(\bm{d'}\) is more successful in distracting the LLM than the original document \(\bm{d}\). Although this measurement does not directly imply generating incorrect answers, we validate the correlation between GPR and the correctness of predictions, as shown in the right panel of Figure~\ref{fig:3}. These results highlight that lowering the GPR tends to induce the generation of more incorrect answers.




\subsection{Process of \textit{GARAG}}

\SetKwInput{KwFunc}{Function}

\begin{algorithm}[t!]
\caption{\small Genetic Attack on RAG}
\label{alg1}
\small
\KwIn{Query $\bm{q}$, Document $\bm{d}$, Number of iterations $N_{\textnormal{iter}}$, Number of parents $N_{\textnormal{parent}}$, Population size $S$, Perturbation rate $pr_{\textnormal{per}}$, Crossover rate $pr_{\textnormal{cross}}$, Mutation rate $pr_{\textnormal{mut}}$}
\KwFunc{Non-dominated sorting $\textnormal{NDS}$, Crowd sorting \textnormal{CS}}
\KwOut{Adversarial document $\bm{d'}^*$}

\tcp{Initialization}
$P_0 \gets \{\bm{d'}_{i}\}_{i=1}^S$ with $pr_{\textnormal{per}}$\;
\For{$i = 1$ \KwTo $N_{\textnormal{iter}}$}{
    \tcp{Crossover}
    $O \gets \textnormal{CROSSOVER}(P_{i-1}, N_{\textnormal{parent}}, pr_{\textnormal{cross}})$\;
    \tcp{Mutation}
    $O \gets \textnormal{MUTATE}(O, pr_{\textnormal{mut}})$\;
    \tcp{Selection}
    $\hat{P}_i \gets P_{i-1} \cup O$\;
    \For{$\bm{d'}$ in $\hat{P}_i$}{
        Evaluate $\mathcal{L}_{\textnormal{RSR}}(\bm{d'})$ and $\mathcal{L}_{\textnormal{GPR}}(\bm{d'})$\;
    }
    $\hat{P}_i \gets \textnormal{CS}{(\textnormal{NDS}(\hat{P}_i))}$\;
    $\bm{d}^* \gets \textnormal{Top-1}(\hat{P}_i)$ \;
    \If{$\bm{a}\neq \texttt{LLM}(\bm{d}^*,\bm{q};\theta) \textbf{ and } \mathcal{L}_{\textnormal{RSR}}(\bm{d}^*) < 1$}{
        \Return $\bm{d}^*$ as adversarial example\;
    }
    $P_i \gets \textnormal{Top-}S(\hat{P}_i)$\;
}
$\bm{d}^* \gets \textnormal{Top-1}(P_{N_{\textnormal{iter}}})$ \;
\Return $\bm{d}^*$ as adversarial example\;
\end{algorithm}
\vspace{-.5em}

The detailed process of \textit{GARAG} is showcased in Algorithm~\ref{alg1}. Our process begins with the initialization of the adversarial document population, and then the population repeats the cycles of crossover, mutation, and selection.

\subsection{Sorting Algorithm}\label{SA}

\begin{algorithm}[t!]
\caption{\small Non-Dominated Sorting Algorithm}
\label{alg2}
\small
    \KwIn{Population $P$}
    \KwOut{Document Set $F_i$ having the front level $i$}
    
    \For{$\bm{d'} \in P$}{
        $S_{\bm{d'}} \gets \emptyset$\;
        $n_{\bm{d'}} \gets 0$\;
        
        \For{$\bm{d''} \in P$}{
            \eIf{$\bm{d'} \prec \bm{d''}$}{
                $S_{\bm{d'}} \gets S_{\bm{d'}} \cup \{\bm{d''}\}$\;
            }
            {\If{$\bm{d''} \prec \bm{d'}$}{
                $n_{\bm{d'}} \gets n_{\bm{d'}} + 1$\;
            
            }}}
        
        \If{$n_{\bm{d'}} = 0$}{
            $\bm{d'}_{\textnormal{rank}} \gets 1$\;
            $F_1 \gets F_1\cup \{\bm{d'}\}$\;
        }
    }
    $i \gets 1$\;
    \While {$F_i \neq \emptyset$}{
        $Q\gets \emptyset$\;
        \For{$\bm{d'} \in F_i$}{
            \For{$\bm{d''} \in S_p$}{
                $n_{\bm{d''}} \gets n_{\bm{d''}} -1$\;
                \If{$n_{\bm{d''}} = 0$}{
                    $\bm{d''}_{\textnormal{rank}} \gets i+1$\;
                    $Q \gets Q \cup \{\bm{d''}\}$\;
                }
            }
        }
        $i \gets i+1$\;
        $F_i \gets Q$\;
    }    
\end{algorithm}
\vspace{-.5em}

In this study, we utilize the sorting algorithms from NSGA-II~\cite{NSGA} to identify the most adversarial documents within extensive search spaces of noisy documents derived from an original document. The algorithm employs non-dominated sorting coupled with crowding distance sorting to organize the population.

\paragraph{Non-Dominated Sorting.} Initially, non-dominated sorting arranges the adversarial documents into different front levels, ensuring that documents within the same level do not dominate one another. The domination relation between the adversarial documents is defined as follows: 

\noindent \textbf{Definition A.1} (Domination). \textit{Given two adversarial documents $\bm{d'}_i$ and $\bm{d'}_j$ perturbed from the original document $\bm{d}$ leading to generate correct answer $\bm{a}$ for a query $\bm{q}$, $\bm{d'}_i$ is said to dominate $\bm{d'}_j$ (i.e., $\bm{d'}_j \prec \bm{d'}_i$) if the following conditions are satisfied:}
\begin{itemize}
    \item $\mathcal{L}_{\textnormal{RSR}}(\bm{d'}_i) < \mathcal{L}_{\textnormal{RSR}}(\bm{d'}_j)$
    \item $\mathcal{L}_{\textnormal{GPR}}(\bm{d'}_i) < \mathcal{L}_{\textnormal{GPR}}(\bm{d'}_j)$
\end{itemize}
The specifics of non-dominated sorting are illustrated in Algorithm~\ref{alg2}.

\paragraph{Crowding Distance Sorting} The crowding distance sorting is applied to rank the documents within each front level. The crowding distance is a crucial part of the algorithm, helping maintain population diversity by giving higher preference to solutions in less crowded regions.

The process of calculating crowding distance in a population begins by assigning each individual a crowding distance value of zero. The population is then sorted in ascending order for each objective function. Boundary points, the first and last individuals in each sorted list, are assigned an infinite crowding distance to ensure their selection. For all other individuals, the crowding distance is calculated by normalizing the difference in objective function values between adjacent individuals, adjusted by the range of the objective values in the population, as given by \(d(i) = d(i) + \frac{f_{i+1} - f_{i-1}}{f_{\text{max}} - f_{\text{min}}}\). This calculation is repeated for each objective function. Finally, the individual crowding distances computed for each objective are summed to estimate the density of solutions surrounding a particular solution, facilitating the selection of diverse solutions in multi-objective optimization.

\subsection{Template}

We adopt the zero-shot prompting template optimal for exact QA tasks, following ~\cite {REAR}, for all LLMs exploited in our experiments.

\begin{tcolorbox}[width=\columnwidth,colback={white},title={\small{QA Template for LLMs}},colbacktitle=white,coltitle=black]
\small{
[INST] Documents:\\
\{Document\}\\
\\
Answer the following question with a very short phrase, such as "1998", "May 16th, 1931", or "James Bond", to meet the criteria of exact match datasets.\\
\\
Question: \{Question\} [/INST]\\
\\
Answer:
}
\end{tcolorbox}

\section{Additional Results}\label{sup:additional_results}

\subsection{Overall Result}

Table~\ref{tab:sup} shows the overall results across three QA datasets, two retrievers, and five LLMs. 

\subsection{Comparison with HotFlip}
\begin{table}[h]
\centering
\small
\caption{\small Comparison with HotFlip Attack on NQ with Contriever and Llama-7b.}
\vspace{-.5em}
\centering
\resizebox{.9\columnwidth}{!}{
\begin{tabular}{l ccc c}
\toprule
 & \multicolumn{3}{c}{\textbf{ASR}} & \textbf{E2E}  \\ \cmidrule(l{2pt}r{2pt}){2-4} \cmidrule(l{2pt}r{2pt}){5-5} 
  & ASR$_R$ & ASR$_L$ & ASR$_T$ & EM \\ \midrule
\textbf{\textit{GARAG}} & 85.9 & 91.1 & 77.5 & 70.1  \\ 
\textbf{\textit{GARAG} on Retriever} & 96.6 & 18.0  & 18.0 & 94.4  \\ 
\textbf{\textit{GARAG} on LLM} & 33.2 & 100.0 & 33.2 & 85.2  \\ \midrule
\textbf{HotFlip on Retriever} & 100.0 & 79.0 & 79.0 & 59.6 \\
\textbf{HotFlip on LLM} & 6.1 & 99.9 & 6.1 & 94.9 \\ \bottomrule
\end{tabular}
}
\label{tab:9-4}
\vspace{-1em}
\end{table}
We compare the vulnerability of low-level perturbations with high-level perturbations implemented by HotFlip~\cite{gradient_2} targeting each module within RAG systems, following the settings of~\citet{PoisonRetriever}.
Note that HotFlip is for high-level perturbations based on word swap, not for low-level perturbations targeting our work.
As shown in Table~\ref{tab:9-4}, HotFlip on the retriever showed a higher attack success rate and significant performance degradation compared to LLM, confirming the retriever acts as a shield for the RAG system. 
Also, HotFlip, with its gradient-based optimization, inevitably finds more adversarial documents than \textit{GARAG}, showing a lower EM score than \textit{GARAG} after the attack. 
However, as ours is the black-box attack just relying on the outputs of the model, not requiring any gradient calculation, it can applied to more diverse scenarios such as exploiting diverse types of perturbations or attacking closed-source models such as ChatGPT~\cite{GPT3}.

\subsection{Adversarial Attack on Closed-source Model}

\begin{table}[h]
\centering
\small
\caption{\small Adversarial attack with \textit{GARAG} on NQ to GPT-3.5}
\vspace{-.5em}
\centering
\resizebox{.9\columnwidth}{!}{
\begin{tabular}{l ccc c}
\toprule
\textbf{Retriever} & \multicolumn{3}{c}{\textbf{ASR}} & \textbf{E2E}  \\ \cmidrule(l{2pt}r{2pt}){2-4} \cmidrule(l{2pt}r{2pt}){5-5} 
  & ASR$_R$ & ASR$_L$ & ASR$_T$ & EM \\ \midrule
\textbf{DPR} & 64.7 & 85.3 & 50.0 & 88.2 \\ 
\textbf{Contriever}  & 74.0 & 86.3 & 60.3 & 83.6 \\ 
\bottomrule
\end{tabular}
}
\label{tab:9-5}
\vspace{-1em}
\end{table}

We further explore the applicability of black-box attacks on the closed-source model, GPT-3.5. Since OpenAI limits access to their models, preventing operations such as gradient calculation for loss objectives, gradient-based attacks like HotFlip~\cite{gradient_2} cannot be applied. However, our proposed method, \textit{GARAG}, can assess the vulnerability of such models as it only requires model outputs for adversarial attacks.
Table~\ref{tab:9-5} presents the results of adversarial attacks on GPT-3.5 with two types of retrievers: DPR and Contriever. Although GPT-3.5 showed some weakness to textual typos, it was more robust than the 7B to 13B size models primarily tested in this experiment. Additionally, the results align with our previous experiments, demonstrating that DPR, which has stronger search performance, is more robust against typos.

\subsection{Changes in Population Distribution Across Iterations in \textit{GARAG}}
\begin{figure}[h]
\centering
\includegraphics[width=0.97\columnwidth]{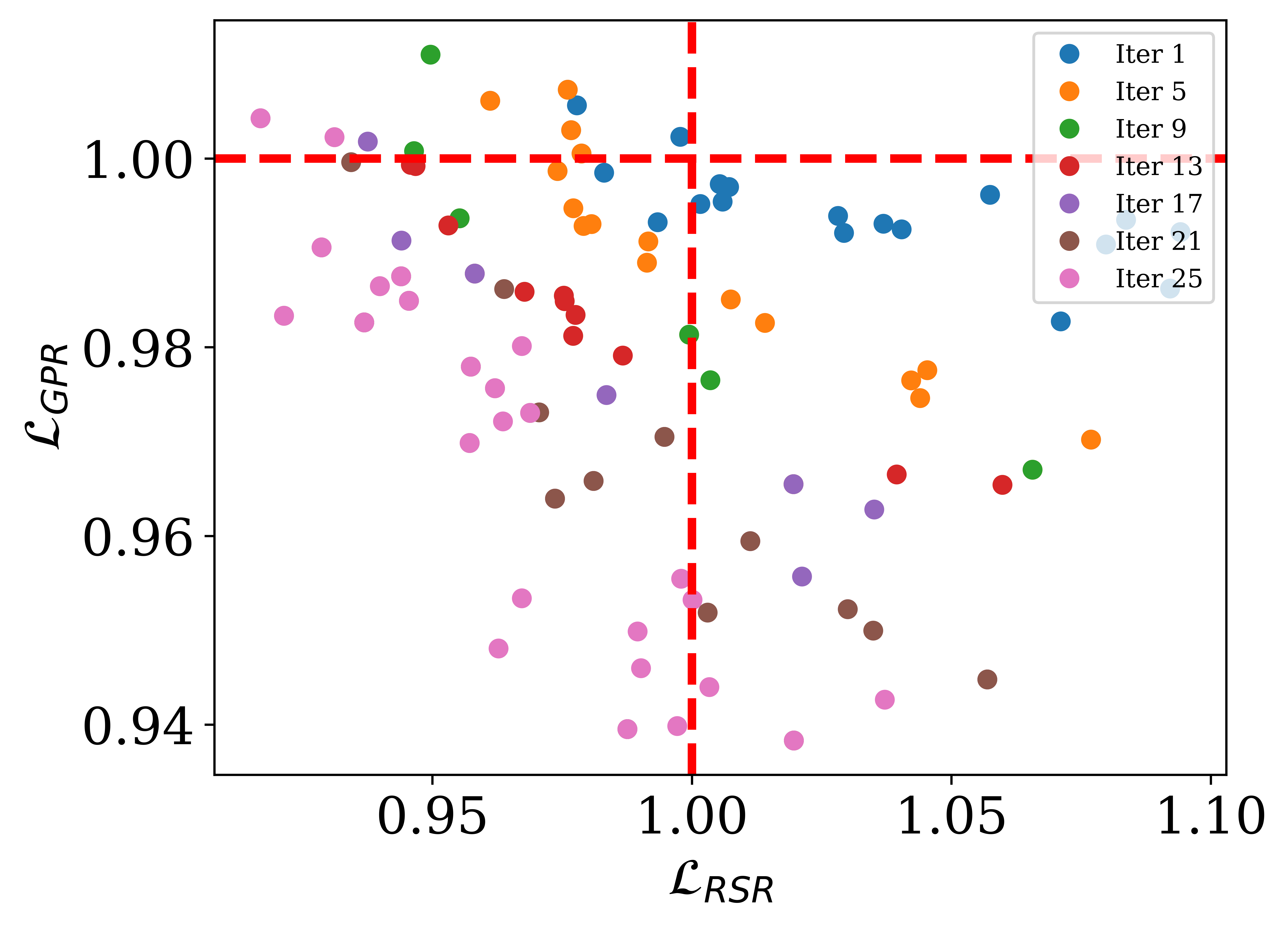}
\caption{\small The process of population refinement by \textit{GARAG} on NQ with Contriever and Llama-7b}
\label{fig:9-4}
\end{figure}
  
We provide a detailed distribution of how the population is refined through the iterative process, as illustrated in Figure~\ref{fig:9-4}. As the iteration number increases, the population distribution progressively converges towards the holistic error zone, demonstrating the effectiveness of \textit{GARAG} in optimization.

\subsection{Case Study}

We conducted case studies with diverse LLMs, including Llama-7b, Vicuna-7b, and Mistral-7b, as shown in Table~\ref{tab:9-2}. 
In all these studies, while the correct answer tokens were not perturbed — allowing for the possibility of grounding correct information — the LLMs typically failed to answer the correct knowledge within the document. 
This often resulted in incorrect predictions or even hallucinations, where the answer was not just wrong but absent from the document. 
However, there was an exception with Mistral-7b, which generated the correct answer and additional explanatory text. While this prediction did not meet the Exact Match (EM) metric, it was semantically correct.

\begin{table*}
\centering
\caption{\small Adversarial attack results of \textit{GARAG} on three QA datasets across different retrievers and LLMs. 
} 
\label{tab:sup}
\resizebox{\textwidth}{!}{
\renewcommand{\arraystretch}{1.0}
\begin{tabular}{ll ccccc ccccc ccccc}
\toprule
& & \multicolumn{5}{c}{\textbf{NQ}} & \multicolumn{5}{c}{\textbf{TriviaQA}} & \multicolumn{5}{c}{\textbf{SQuAD}} \\ \cmidrule(l{2pt}r{2pt}){3-7} \cmidrule(l{2pt}r{2pt}){8-12} \cmidrule(l{2pt}r{2pt}){13-17} 
& & \multicolumn{3}{c}{\textbf{ASR($\uparrow$)}} & \multicolumn{2}{c}{\textbf{E2E($\downarrow$)}} & \multicolumn{3}{c}{\textbf{ASR($\uparrow$)}} & \multicolumn{2}{c}{\textbf{E2E($\downarrow$)}} & \multicolumn{3}{c}{\textbf{ASR($\uparrow$)}} & \multicolumn{2}{c}{\textbf{E2E($\downarrow$)}} \\
\textbf{Retriever} & \textbf{LLM} & ASR$_R$ & ASR$_L$ & ASR$_T$ &  EM & Acc. & ASR$_R$ & ASR$_L$ & ASR$_T$ &  EM & Acc. & ASR$_R$ & ASR$_L$ & ASR$_T$ &  EM & Acc.  \\ \midrule
\multirow{6}*{\textbf{DPR}} & \textbf{Llama2-7b}  & 75.4 & 89.8 & 66.0 & 76.8 & 80.6 & 78.2 & 91.7 & 70.2 & 81.6 & 85.3& 84.1 & 90.1 & 74.2 & 73.0 & 78.  \\
& \textbf{Llama2-13b} & 71.3 & 91.7 & 63.5 & 82.8 & 88.2 & 83.9 & 92.0 & 76.1 & 76.7 & 83.3 & 80.0 & 92.4 & 72.7 & 86.3 & 90.5  \\ \noalign{\vskip 0.5ex}\cdashline{2-17}\noalign{\vskip 0.5ex}
                   & \textbf{Vicuna-7b}  & 83.0 & 81.6 & 65.1 &  62.0 & 79.2 & 91.1 & 79.5 & 70.8 & 58.4 & 81.7 & 92.0 & 81.1 & 73.4 & 51.2 & 76.9  \\
& \textbf{Vicuna-13b} & 82.8 & 80.9 & 64.4 & 58.5 & 83.3 & 91.8 & 83.5 & 75.4 & 59.2 & 85.7 & 91.7 & 80.5 & 72.5 & 57.4 & 80.5  \\
\noalign{\vskip 0.5ex}\cdashline{2-17}\noalign{\vskip 0.5ex}
                   & \textbf{Mistral-7b}  & 78.5 & 85.9 & 65.1 & 69.1 & 96.5 & 84.7 & 84.9 & 69.8 & 66.5 & 97.7 & 87.8 & 85.7 & 73.5 & 64.4 & 95.2  \\ \midrule \midrule
\multirow{6}*{\textbf{Contriever}} & \textbf{Llama2-7b} & 85.9 & 91.1 & 77.5 & 70.1 & 74.7 & 84.9 & 90.7 & 76.0 & 82.0 & 86.9 & 85.2 & 91.2 & 76.4 & 72.9 & 77.2  \\
& \textbf{Llama2-13b} & 78.9 & 91.2 & 70.5 & 78.7 & 85.7 & 81.0 & 91.9 & 72.9 & 86.2 & 91.7 & 86.1 & 93.0 & 79.1 & 77.2 & 84.5 \\ \noalign{\vskip 0.5ex}\cdashline{2-17}\noalign{\vskip 0.5ex}
                   & \textbf{Vicuna-7b} & 90.8 & 81.3 & 72.4 & 52.2 & 72.5 & 93.0 & 80.8 & 74.0 & 60.3 & 81.5 & 92.6 & 82.5 & 75.2 & 52.7 & 76.7 \\
& \textbf{Vicuna-13b} & 87.5 & 85.5 & 73.3 & 63.9 & 95.4 & 88.8 & 86.4 & 75.2 & 66.2 & 97.8 & 91.2 & 88.0 & 79.3 & 59.2 & 92.6  \\ \noalign{\vskip 0.5ex}\cdashline{2-17}\noalign{\vskip 0.5ex}
                   & \textbf{Mistral-7b} & 87.5 & 85.5 & 73.3 & 63.9 & 95.4 & 88.8 & 86.4 & 75.2 & 66.2 & 97.8 & 91.2 & 88.0 & 79.3 & 59.2 & 92.6  \\ \bottomrule

\end{tabular}
}
\end{table*}


\begin{table*}[t!]
\vspace{-.5em}
\caption{\small Case study on NQ with Contriever, where perturbed texts are in\hlc[red!25]{red} and correct answers are in\hlc[blue!25]{blue}.}
\label{tab:9-2}
\vspace{-.5em}
\small
\centering
\resizebox{\textwidth}{!}{
\renewcommand{\arraystretch}{0.8}
\begin{tabular}{cc}
\toprule 
\multicolumn{2}{c}{\textit{Llama-7b}} \\ \midrule 
\multicolumn{1}{p{.15\textwidth}}{\textbf{Question}} & \multicolumn{1}{p{.85\textwidth}}{Which site of an enzyme is called allosteric site?} \\  \noalign{\vskip 0.5ex}\cdashline{1-2}\noalign{\vskip 0.5ex}
\multicolumn{1}{p{.15\textwidth}}{\textbf{Noisy Document}} & \multicolumn{1}{p{.85\textwidth}}{\hlc[red!25]{A;losteric} enzyme Long-range allostery is \hlc[red!25]{esprcially} \hlc[red!25]{imponant} in cell signaling. Allosteric regulation is also particularly important in the cell's \hlc[red!25]{abil9ty} to \hlc[red!25]{adjusy} enzyme activity. The term "allostery" comes from the Greek "allos", "other," and "stereos", "\hlc[red!25]{silid} (object)." This is in reference to the fact that the \hlc[blue!25]{regulatory site} of an allosteric protein is physically distinct from its active site. The protein catalyst (enzyme) may be \hlc[red!25]{paft} of a multi-subunit complex, and/or may transiently or permanently \hlc[red!25]{associatr} with a Cofactor (e.g. adenosine triphosphate). Catalysis of \hlc[red!25]{biochejical} reactions is vital due to the very \hlc[red!25]{law} reaction rates of the uncatalysed \hlc[red!25]{reactioms}.}  \\ \midrule
\multicolumn{1}{p{.15\textwidth}}{\textbf{Answer}} & \multicolumn{1}{p{.85\textwidth}}{Regulatory site} \\ \noalign{\vskip 0.5ex}\cdashline{1-2}\noalign{\vskip 0.5ex}
\multicolumn{1}{p{.15\textwidth}}{\textbf{Prediction}} & \multicolumn{1}{p{.85\textwidth}}{Active site} \\ \midrule \midrule

\multicolumn{1}{p{.15\textwidth}}{\textbf{Question}} & \multicolumn{1}{p{.85\textwidth}}{Who did Cora marry in once upon a time?} \\  \noalign{\vskip 0.5ex}\cdashline{1-2}\noalign{\vskip 0.5ex}
\multicolumn{1}{p{.15\textwidth}}{\textbf{Noisy Document}} & \multicolumn{1}{p{.85\textwidth}}{The Miller\'s Daughter (Once Upon a Time) to the King and accepts \hlc[blue!25]{Henry}'s resultant marriage proposal. The day before her wedding, Cora \hlc[red!25]{ques6ions} her \hlc[red!25]{olans}; she is unlikely to become \hlc[red!25]{qjeen} as Henry is fifth in \hlc[red!25]{linf} to the throne, while Rumplestiltskin, with whom she has been having an affair, offers her love. They agree to amend the \hlc[red!25]{contratc} so Cora owes Rumplestiltskin "his" child. He also agrees to teach her how to take a heart, so that she can \hlc[red!25]{kilk} King Savier. That night, she \hlc[red!25]{confromts} the king. He reveals that he knows of her relationship with Rumplestiltskin; telling her that "\hlc[red!25]{pove} is weakness," he \hlc[red!25]{ays}}  \\ \midrule
\multicolumn{1}{p{.15\textwidth}}{\textbf{Answer}} & \multicolumn{1}{p{.85\textwidth}}{Henry} \\ \noalign{\vskip 0.5ex}\cdashline{1-2}\noalign{\vskip 0.5ex}
\multicolumn{1}{p{.15\textwidth}}{\textbf{Prediction}} & \multicolumn{1}{p{.85\textwidth}}{Rumplestiltskin} \\ \midrule \midrule

\multicolumn{2}{c}{\textit{Vicuna-7b}} \\ \midrule 
\multicolumn{1}{p{.15\textwidth}}{\textbf{Question}} & \multicolumn{1}{p{.85\textwidth}}{What is the 3rd largest state in USA?} \\  \noalign{\vskip 0.5ex}\cdashline{1-2}\noalign{\vskip 0.5ex}
\multicolumn{1}{p{.15\textwidth}}{\textbf{Noisy Document}} & \multicolumn{1}{p{.85\textwidth}}{\hlc[red!25]{Wextern} United States LGBT community, and Oakland, \hlc[blue!25]{California} has a large \hlc[red!25]{percen5age} of residents being African-American, as well as Long \hlc[red!25]{Beadh}, \hlc[blue!25]{California} which also has a large Black community. \hlc[red!25]{Ths} state of Utah has a Mormon majority (estimate at 62.4\% in 2004), while some cities like Albuquerque, \hlc[red!25]{Nrw} Mexico; \hlc[red!25]{Billkngs}, \hlc[red!25]{Montqna}; Spokane, Washington; and Tucson, Arizona are located near Indian Reservations. In remote areas there are settlements of \hlc[blue!25]{Alaskan} Natives and Native Hawaiians. \hlc[red!25]{Fqcing} both the \hlc[red!25]{Pacitic} Ocean and the Mexican border, the West has been shaped by a \hlc[red!25]{cariety} of ethnic groups. Hawaii is the only state in the union in which}  \\ \midrule
\multicolumn{1}{p{.15\textwidth}}{\textbf{Answer}} & \multicolumn{1}{p{.85\textwidth}}{California, Alaska} \\ \noalign{\vskip 0.5ex}\cdashline{1-2}\noalign{\vskip 0.5ex}
\multicolumn{1}{p{.15\textwidth}}{\textbf{Prediction}} & \multicolumn{1}{p{.85\textwidth}}{Oregon} \\ \midrule \midrule

\multicolumn{1}{p{.15\textwidth}}{\textbf{Question}} & \multicolumn{1}{p{.85\textwidth}}{When did the movie peter pan come out} \\  \noalign{\vskip 0.5ex}\cdashline{1-2}\noalign{\vskip 0.5ex}
\multicolumn{1}{p{.15\textwidth}}{\textbf{Noisy Document}} & \multicolumn{1}{p{.85\textwidth}}{Peter Pan as an adult by Robin Williams, with \hlc[red!25]{iblue} eyes and dark brown hair; in flashbacks to him in his youth, his \hlc[red!25]{hwir} is light brown. In this film his ears \hlc[red!25]{ap;ear} \hlc[red!25]{poin} only when he is \hlc[red!25]{Petef} Pan, not as Peter Banning. His Pan attire resembles the \hlc[red!25]{D9sney} outfit (minus the cap). In the \hlc[red!25]{lvie-action} \hlc[blue!25]{2003} "\hlc[red!25]{-eter} Pan" film, he is portrayed by Jeremy \hlc[red!25]{Su,pter}, who has blond hair and blue-green eyes. His outfit is made of leaves and vines. J.M. Barrie created his character based on his older brother, \hlc[red!25]{Davic}, who died in an ice-skating \hlc[red!25]{accieent} the day before}  \\ \midrule
\multicolumn{1}{p{.15\textwidth}}{\textbf{Answer}} & \multicolumn{1}{p{.85\textwidth}}{25 December 2003, 2003} \\ \noalign{\vskip 0.5ex}\cdashline{1-2}\noalign{\vskip 0.5ex}
\multicolumn{1}{p{.15\textwidth}}{\textbf{Prediction}} & \multicolumn{1}{p{.85\textwidth}}{1998} \\ \midrule \midrule

\multicolumn{2}{c}{\textit{Mistral-7b}} \\ \midrule 
\multicolumn{1}{p{.15\textwidth}}{\textbf{Question}} & \multicolumn{1}{p{.85\textwidth}}{Make it or break it who goes to the Olympics?} \\  \noalign{\vskip 0.5ex}\cdashline{1-2}\noalign{\vskip 0.5ex}
\multicolumn{1}{p{.15\textwidth}}{\textbf{Noisy Document}} & \multicolumn{1}{p{.85\textwidth}}{Make It or Break It A new gymnast, Max (Josh Bowman), \hlc[red!25]{cpmes} to The Rock, \hlc[red!25]{wttracting} the attention of \hlc[blue!25]{Lauren} and \hlc[blue!25]{Payson}. Though Max seems more interested in \hlc[blue!25]{Payson}, she is more focused on her dream. \hlc[blue!25]{Lauren} tells \hlc[blue!25]{Payson} that Max is her \hlc[red!25]{nww} boyfriend; Max denies this. Mas, Austin, and \hlc[blue!25]{Payson} hope \hlc[blue!25]{Kaylie} train in secret, as they try to convince Sasha that \hlc[blue!25]{Kaylie} is ready to compete at Worlds. In the week of Emily's departure, Kelly Parke becomes closer to \hlc[blue!25]{Kaylie}, \hlc[red!25]{des-ite} the scheming of her stage her, Sheila (Kathy Najimy). \hlc[blue!25]{Kaylie} offers her comfort as Kelly's mom is only \hlc[red!25]{sup[ortive} of Kelly's}  \\ \midrule
\multicolumn{1}{p{.15\textwidth}}{\textbf{Answer}} & \multicolumn{1}{p{.85\textwidth}}{Jordan, Payson, Lauren, Colleen, Kaylie} \\ \noalign{\vskip 0.5ex}\cdashline{1-2}\noalign{\vskip 0.5ex}
\multicolumn{1}{p{.15\textwidth}}{\textbf{Prediction}} & \multicolumn{1}{p{.85\textwidth}}{Kaylie goes to the Olympics} \\ \midrule \midrule

\multicolumn{1}{p{.15\textwidth}}{\textbf{Question}} & \multicolumn{1}{p{.85\textwidth}}{When did Michael Jordan get his last ring?} \\  \noalign{\vskip 0.5ex}\cdashline{1-2}\noalign{\vskip 0.5ex}
\multicolumn{1}{p{.15\textwidth}}{\textbf{Noisy Document}} & \multicolumn{1}{p{.85\textwidth}}{\hlc[red!25]{Hisfory} of basketball A large place of this is due to the transcendent stars that have played the game through the years. It was because of the play of Michael Jordan that basketball started to \hlc[red!25]{rsach} international audiences, especially on the 992 United States men's Olympic basketball team, known as the Dream Team. After his \hlc[red!25]{fianl} championship and second retirement in \hlc[blue!25]{1998}, there was a void as in who would be the face of basketball. So;n after with the help of Shaquille O'Neal, \hlc[red!25]{Koge} Bryant would go on to win three straight championships from 2000–2002 with the Los Angeles \hlc[red!25]{kers}, helping make basketball}  \\ \midrule
\multicolumn{1}{p{.15\textwidth}}{\textbf{Answer}} & \multicolumn{1}{p{.85\textwidth}}{1998} \\ \noalign{\vskip 0.5ex}\cdashline{1-2}\noalign{\vskip 0.5ex}
\multicolumn{1}{p{.15\textwidth}}{\textbf{Prediction}} & \multicolumn{1}{p{.85\textwidth}}{2002. (This is based on the context of the given document stating that Kobe Bryant and Shaquille O'Neal)} \\

\bottomrule

\end{tabular}
}
\label{tab:case}
\end{table*}

\end{document}